\documentclass[fleqn,10pt]{wlscirep}
\usepackage[utf8]{inputenc}
\usepackage[T1]{fontenc}
\usepackage{caption}
\usepackage{subcaption}
\usepackage{amsmath}
\usepackage{amsfonts}
\usepackage{amssymb}
\usepackage[flushleft]{threeparttable}
\usepackage{multirow}
\usepackage{tablefootnote}
\usepackage{array,makecell}
\title{Pulmonologists-Level lung cancer detection based on standard blood test results and smoking status using an explainable machine learning approach}  

\author[1,+]{Ricco Noel Hansen Flyckt}
\author[1,+]{Louise Sjodsholm}
\author[2,+]{Margrethe Høstgaard Bang Henriksen}
\author[3,5]{Claus Lohman Brasen}
\author[1]{Ali Ebrahimi}
\author[4,5]{Ole Hilberg}
\author[2,5]{Torben Frøstrup Hansen}
\author[1]{Uffe Kock Wiil}
\author[2]{Lars Henrik Jensen}
\author[1,*]{Abdolrahman Peimankar}
\affil[1]{SDU Health Informatics and Technology, The M\ae rsk Mc-Kinney M\o ller Institute, University of Southern Denmark, 5230 Odense, Denmark}
\affil[2]{Department of Oncology, Vejle Hospital, University Hospital of Southern Denmark, 7100 Vejle, Denmark}
\affil[3]{Department of Biochemistry and Immunology, Vejle Hospital, University Hospital of Southern Denmark, 7100 Vejle, Denmark}
\affil[4]{Department of Internal Medicine, Vejle Hospital, University Hospital of Southern Denmark, 7100 Vejle, Denmark}
\affil[5]{Institute of Regional Health Research, University of Southern Denmark, 5230 Odense, Denmark}

\affil[*]{abpe@mmmi.sdu.dk}
\affil[+]{these authors contributed equally to this work}

\begin{abstract}
Lung cancer (LC) remains the primary cause of cancer-related mortality, largely due to late-stage diagnoses. Effective strategies for early detection are therefore of paramount importance. In recent years, machine learning (ML) has demonstrated considerable potential in healthcare by facilitating the detection of various diseases. In this retrospective development and validation study, we developed an ML model based on dynamic ensemble selection (DES) for LC detection. The model leverages standard blood sample analysis and smoking history data from a large population at risk in Denmark. The study includes all patients examined on suspicion of LC in the Region of Southern Denmark from 2009 to 2018. We validated and compared the predictions by the DES model with diagnoses provided by five pulmonologists. Among the 38,944 patients, 9,940 had complete data of which 2,505 (25\%) had LC. The DES model achieved an area under the roc curve of 0.77$\pm$0.01, sensitivity of 76.2\%$\pm$2.4\%, specificity of 63.8\%$\pm$2.3\%, positive predictive value of 41.6\%$\pm$1.2\%, and F\textsubscript{1}-score of 53.8\%$\pm$1.1\%. The DES model outperformed all five pulmonologists, achieving a sensitivity 9\% higher than their average. The model identified smoking status, age, total calcium levels, neutrophil count, and lactate dehydrogenase as the most important factors for the detection of LC. The results highlight the successful application of the ML approach in detecting LC, surpassing pulmonologists' performance. Incorporating clinical and laboratory data in future risk assessment models can improve decision-making and facilitate timely referrals.

\end{abstract}
\begin{document}

\flushbottom
\maketitle

\thispagestyle{empty}

\section*{Introduction} \label{intro}
Lung cancer (LC) is the leading cause of cancer-related deaths, and ranks as the second most prevalent cancer type globally, with 2,21 million new cases in 2020 \cite{sharma2022mapping,sung2021global}. While survival rates have seen improvements over the past decade, one-year survival still remains low \cite{jakobsen2016mortality,esund}. Late-stage diagnosis limits the possibility of curative treatment and early referral for diagnostics is therefore crucial to reduce the growing healthcare burden \cite{Annualreport}.

Several countries have introduced screening of LC among high-risk individuals based on the American National Lung Screening Trial (NLST) and the Dutch/Belgian randomized LC screening trial (NELSON). They demonstrated a reduction in mortality up to 25\% depending on screening method \cite{smith2019cancer,aberle2011reduced,dawson2020nelson}. Despite these promising results, there is an argument for the integration of additional risk factors into prediction models, to improve sensitivity and cost-effectiveness \cite{lam2021contemporary,liu2020evolving,de2020reduced}.

The interest in detecting LC through liquid biopsies containing circulating tumor DNA and additional biomarkers have been increasing, but the lack of standardization has hindered the implementation of the approach \cite{di2021liquid}. Routine blood tests, although more convenient, efficient, and affordable, have seen limited use in predicting LC \cite{gould2021machine,wang2019prediction}. Previous studies have achieved positive results in detecting and predicting LC based on routine blood tests, but the models were based on unrepresentative cohorts and relied on imputation of large amounts of missing data. 

\begin{figure}[ht]
    \centering
    \includegraphics[width=0.95\linewidth]{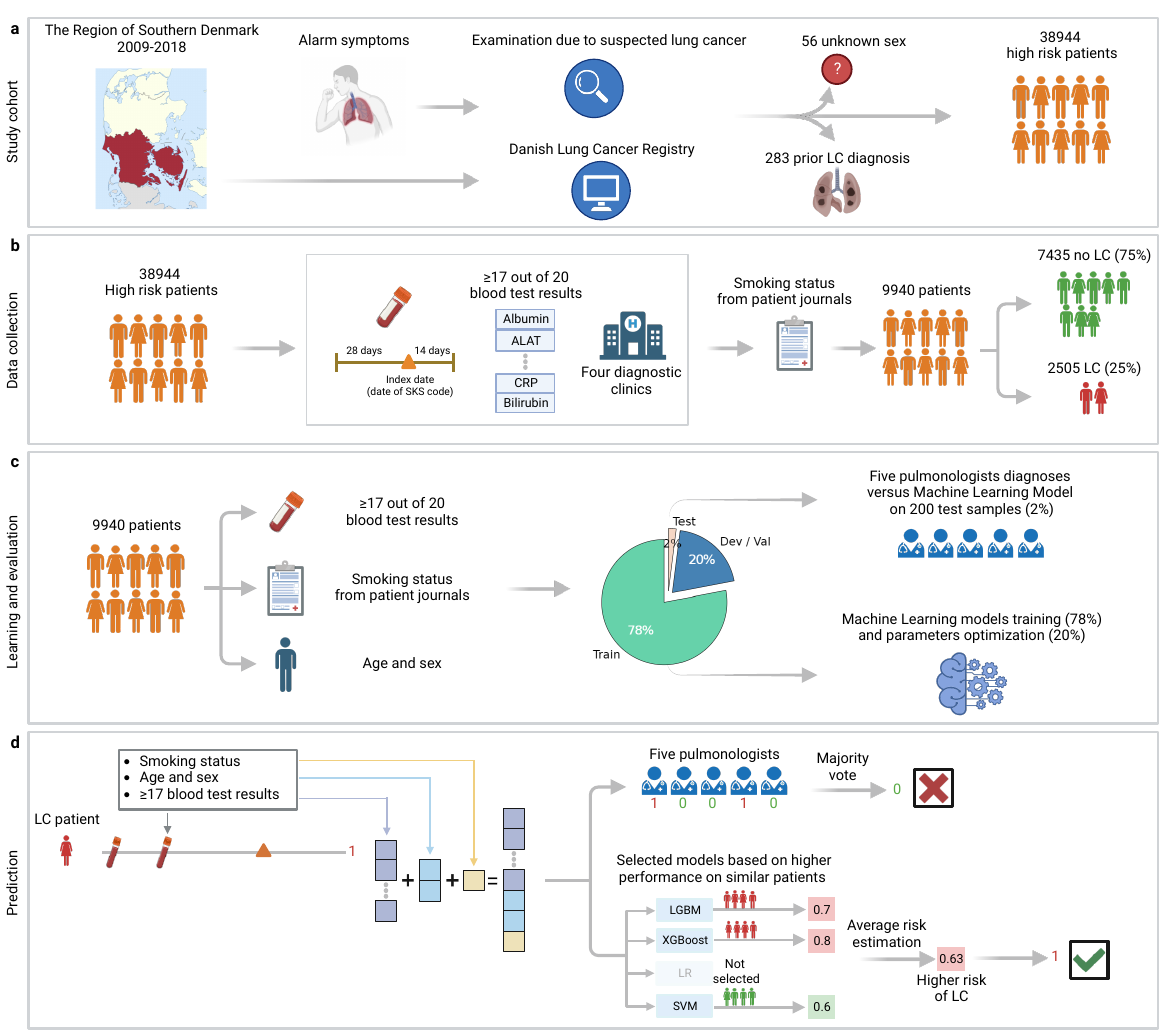}
    \caption{Flowchart illustrating the LC detection from laboratory and smoking status data. (\textbf{a}) The composition of the study cohort. (\textbf{b}) The inclusion criteria for the data collection of patients who were suspicious of having LC. (\textbf{c}) The workflow of splitting the data into train, validation, and test sets. The train and validation sets are used for the learning process of the model and to minimize the prediction/detection error. The test set of 200 samples are utilized for the comparison between the model’s prediction and five pulmonologists diagnosis. (\textbf{d}) The collected data from different sources are concatenated to be used as inputs for the DES model and to be also provided for the pulmonologists in a fair manner for their diagnoses.}
    \label{fig:fig1}
\end{figure}

This study retrospectively collected data from all patients in the Region of Southern Denmark referred for examination on suspicion of LC between January 2009 and December 2018 \cite{henriksen2023collection}. In this study our objective was to compare the performance of various machine learning (ML) models in detecting LC patients. Our approach, which relied exclusively on smoking status, age, gender, and routine blood test results to predict LC, facilitated straightforward integration into clinical settings through an ensemble-based ML model. Additionally, we validated our proposed model by comparing its diagnostic performance with the diagnoses provided by five pulmonologists in a subset of 200 cases. The results are presented using explainable modules designed to assist clinicians in interpreting the model’s predictions. Figure \ref{fig:fig1} gives an overview of the study cohort, data collection and methodologies applied in this study.

\section*{Results}\label{sec:Results}
\subsection*{Demographic and baseline clinical characteristics of patients}
A total of 9,940 patients met the inclusion criteria, of which 2,505 (25\%) had LC and 7,435 (75\%) did not. The median age of the LC patients was 74 years (IQR 68-80), and 71 years in the non-LC patients (IQR 59-79). The LC group consisted of 52\% females in contrast to 44\% in the non-LC group. Approximately 92\% of LC patients were either current or former smokers, whereas the proportion was 69\% among non-LC patients. Table 1 detail clinical variables and blood test results.

\subsection*{Prediction performance of LC detection models}
\begin{figure}[ht]
    \centering
    \includegraphics[width=0.7\linewidth]{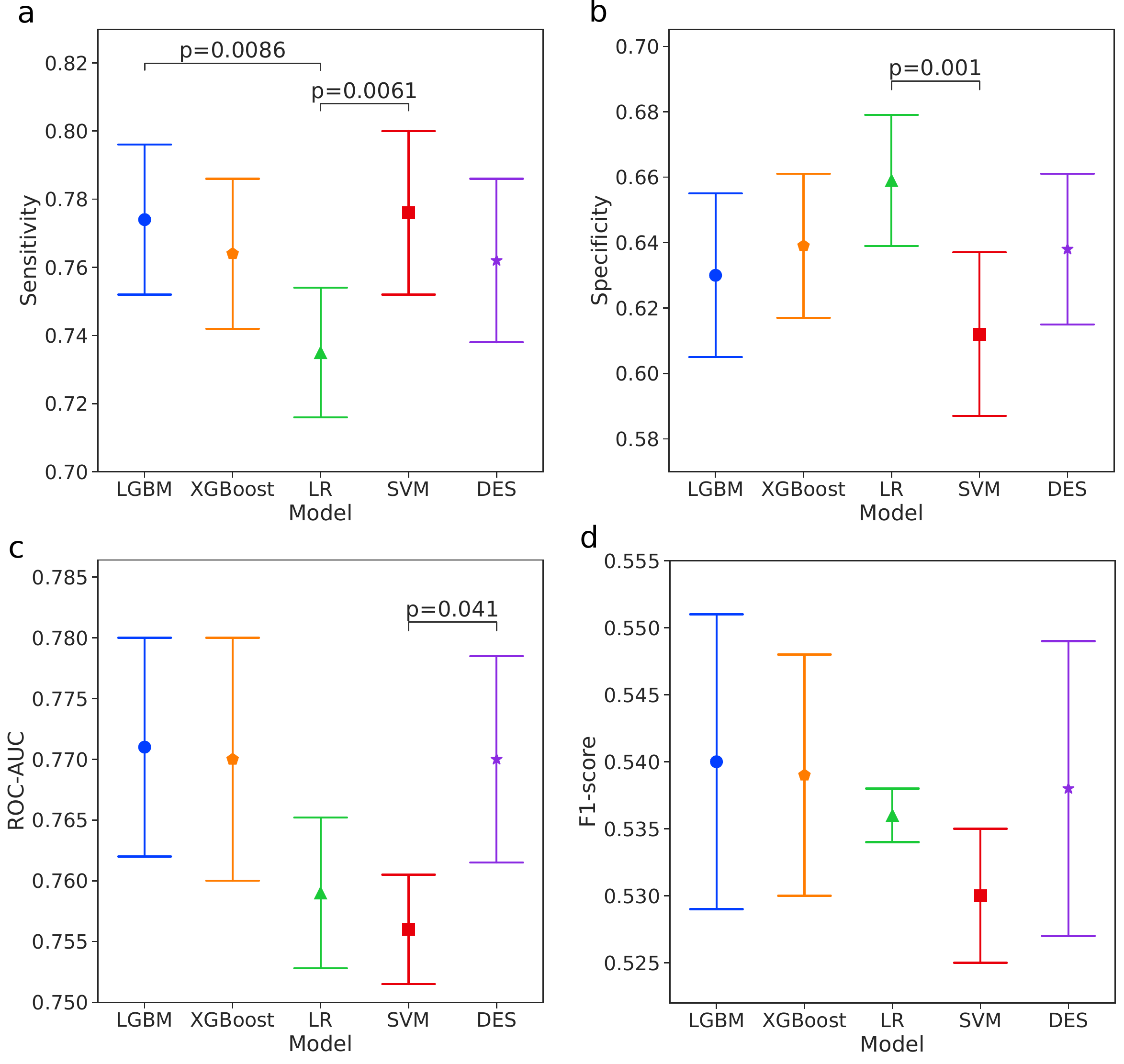}
    \caption{Comparison of evaluation metrics for the validation set using 5-fold cross-validation. (\textbf{a}) Models comparison using sensitivity metric. There is a significant difference between the two highest models (i.e., LGBM and SVM) and LR. (\textbf{b}) Models comparison using specificity metric. There is only significant difference between DES and LR. (\textbf{c}) Models comparison using ROC-AUC metric. There is only significant difference between DES and SVM. (\textbf{d}) Models comparison using F\textsubscript{1}-score metric. There is no significant difference between the models. The central marker represents mean values along with corresponding standard deviations. The horizontal brackets indicate significant differences in performance, as determined by the Nemenyi post-hoc test, with a two-sided p-value threshold of 0.05.}
    \label{fig:fig2}
\end{figure}

Figure \ref{fig:fig2} shows the mean and standard deviation of classification performances of all the models in the validation set using 5-fold cross-validation. The SVM demonstrated the highest median sensitivity, yet it did not exhibit a statistically significant difference when compared to the LGBM, XGBoost, and DES classifiers (Fig. \ref{fig:fig2}a). Conversely, the LGBM classifier achieved the highest median ROC-AUC, but this result was not statistically significant when compared to the other four models (Fig. \ref{fig:fig2}c). The Nemenyi Post-hoc test conclusively showed that no model consistently outshone the others. Consequently, we chose the ensemble model (DES), which combines all four classification models (LGBM, XGBoost, LR, and SVM). The ensemble model ensures enhanced generalizability when applied to new samples within clinical settings.

\subsection*{Explainable LC prediction performance}
\begin{figure}[ht]
    \centering
    \includegraphics[width=0.95\linewidth]{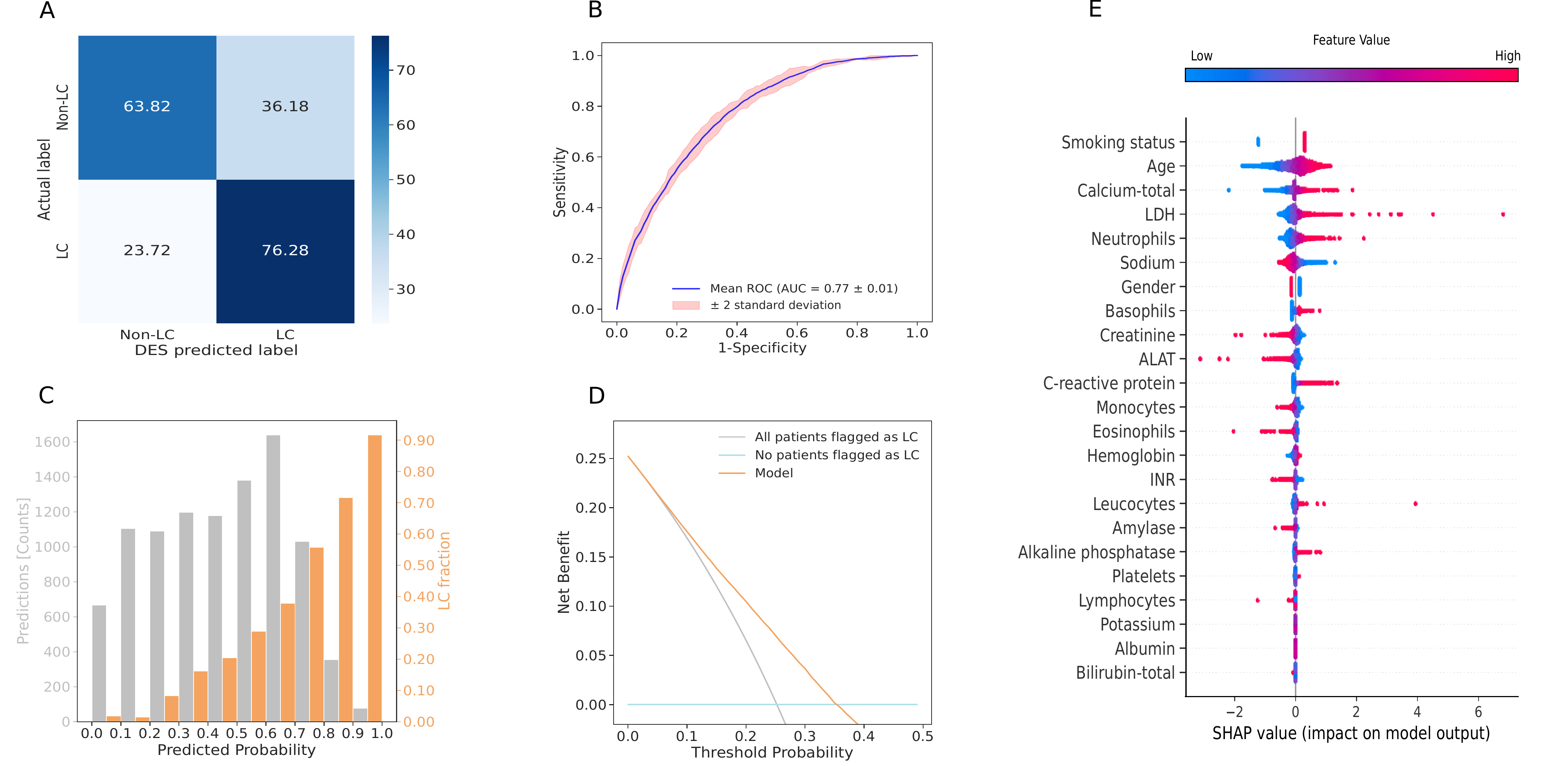}
    \caption{Assessment of the Dynamic Ensemble Selection Model (DES) through 5-fold cross-validation. (\textbf{a}) Average confusion matrix for 5-fold cross-validation. (\textbf{b}) Average ROC curve for 5-fold cross-validation. The highlighted pink area around the ROC curve represents the standard deviation of 5-fold cross-validation. (\textbf{c}) Predicted probabilities compared to observed LC cases showing the number of patients on the left y-axis and the fraction of patients on the right y-axis. Predicted probabilities are categorized into bins of 0.1. For instance, in the range of 0.7-0.8 (70-80\%), the actual fraction of LC cases were 0.55 (55\%), corresponding to 1000 patients with LC out of the total cases. (\textbf{d}) Decision curve analyses displaying the relationship between threshold probablilities and the net benefit when utilizing the DES-model for classification of patients at high risk of LC. This is compared to selecting all patients (grey line) or no patients (blue line). The DES-model demonstrates a higher net benefit across threshold probabilities ranging from approximately 7\% to 35\% compared to the other two clinical strategies. (\textbf{e}) SHAP summary plot with features listed in descending order of importance.}
    \label{fig:fig3}
\end{figure}

At a default risk-threshold of 0.5, the DES algorithm correctly identifies 76.28\% of the LC patients and 63.82\% of the non-LC patients. However, it exhibits a false-positive rate of 36.18\% (Fig. \ref{fig:fig3}a). The DES model achieves a mean ROC-AUC of 0.77±0.01 (Fig. \ref{fig:fig3}b), and the low standard deviation underscores the model's stability during 5-fold cross-validated evaluations.

Figure \ref{fig:fig4}c illustrates the distribution of predicted probabilities versus the actual LC incidence within each interval. LC incidence consistently rises with increasing probability across all intervals, but there is a systematic overestimation of the predicted risk. For instance, among patients with an estimated mean predicted probability between 0.4 and 0.5, the true fraction of LC patients is only 0.2.

The decision-curve analyses are depicted in Fig. \ref{fig:fig3}d, providing insight into clinical utility at different threshold probabilities. The analyses reveal that below a threshold probability of approximately 7\%, there is no distinction between flagging all patients as LC cases and using the model to discern LC cases. Conversely, above a threshold probability of around 7\% the net benefit increases for the model, indicating greater clinical usefulness compared to flagging all patients as LC cases. At a probability of approximately 35\% the model's net benefit equals that of not flagging any patients as LC cases. Hence, the model outperforms the other two clinical strategies for threshold probabilities ranging from around 7\% to 35\%.

Figure \ref{fig:fig3}e presents a summary plot employing SHAP values displaying the most critical input features for LC detection. Active or current smoking status, advanced age, elevated levels of total calcium, LDH, and neutrophil count, as well as low values of sodium and female gender, are the eight most important features. Post hoc analyses demonstrate that the model's performance remains consistent when limited to these eight features (see Supplementary Fig. S10 online). To provide detailed insight into the models' decisions for individual cases, SHAP values were employed to interpret the predictions for every patient (see Supplementary Figs. S11-14 online).

\subsection*{Pulmonologists-level LC prediction}
\begin{figure}[ht]
    \centering
    \includegraphics[width=0.7\linewidth]{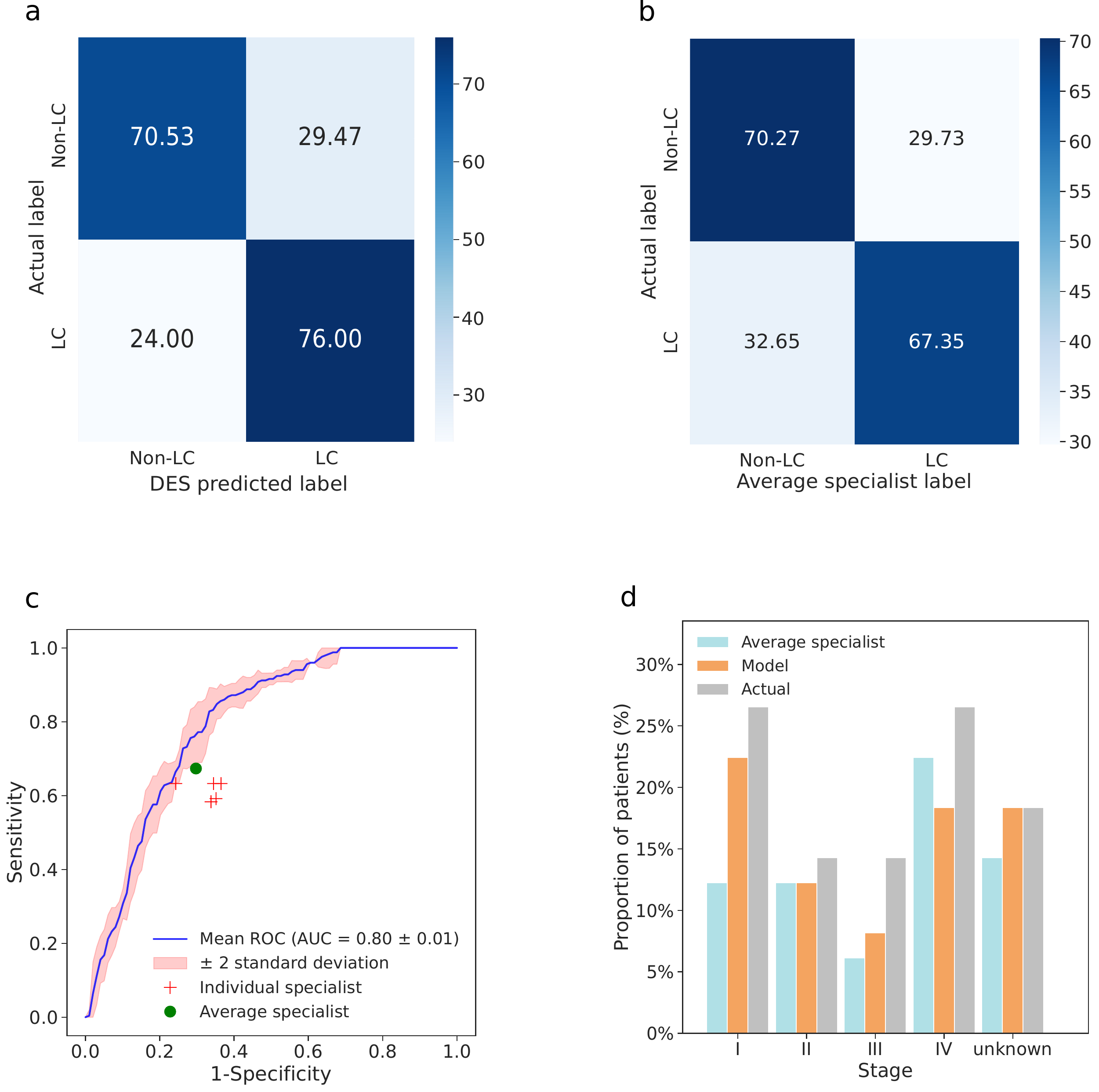}
    \caption{Assessment of the DES model on the 200 samples and the comparison with pulmonologists. (\textbf{a}) Confusion matrix representing the DES model’s prediction versus the actual diagnosis. (\textbf{b}) Confusion matrix of the predictions made by the averaged pulmonologists votes versus the actual diagnosis. (\textbf{c}) ROC curve with the individual pulmonologist’s performance marked by red marks and averaged performance marked by a green dot. (\textbf{d}) Correct predictions of the DES model and averaged pulmonologists in relation to the four stages of lung cancer, alongside the actual distribution of each stage.}
    \label{fig:fig4}
\end{figure}

The performance of the classification algorithms was compared with the diagnoses made by five pulmonologists using 200 hold out samples (see Supplementary Table 4 online). Notably, the LGBM model appeared as the top-performing classifier, achieving accuracy, sensitivity, positive predictive value and F1-score of 73.3\%, 77.2\%, 48.6\%, and 58.9\%, respectively. Overall, all models demonstrated comparable performance across various metrics, with the Dynamic Ensemble Selection (DES) model recognized as the most robust classifier. The averaged pulmonologists' diagnoses attained a sensitivity of 67.4\% and a specificity of 70.3\% (Fig. \ref{fig:fig4}a). At the same level of specificity, the DES model exhibited superior sensitivity, reaching 76.0\% on the same 200 patients (Fig. \ref{fig:fig4}b). This represents a significant improvement, determined by the Nemenyi Post-hoc test, of more than 8\% points over the pulmonologists' performance (p=0.002). Figure \ref{fig:fig4}c presents the ROC curve for the DES model applied to the 200 samples, along with the individual and averaged performance of the pulmonologists. In Fig. \ref{fig:fig4}d, the distribution of actual LC patients in each stage is compared to the correctly predicted patients by both the model and pulmonologists on the 200 samples. The analysis shows that, on average, specialists excel in diagnosing patients with stage IV of LC, while the model outperforms specialists in stages I and III. The model closely aligns with the actual number of patients in stage II.

\section*{Discussion}
In this study we developed a classification model using data from 9,940 high-risk individuals who had undergone examinations on suspicion of LC in the Region of Southern Denmark. The final model was constructed as an ensemble model (DES), leveraging the strengths of four established ML models. The DES model exhibited the ability to classify LC patients with an ROC-AUC of 0.77 on the validation set. Five pulmonologists independently evaluated 200 samples, achieving a sensitivity of 67.4\% and a specificity of 70.3\%. When matched for specificity, the DES model surpassed the pulmonologists by 8\% points. The SHAP summary plot identified the top eight influential features, including active or former smoker status, advanced age, elevated levels of total calcium, neutrophil count, and LDH, as well as reduced levels of sodium and female gender. Using only these eight features for model training yielded a performance equivalent to using all available features. Consequently, these eight factors can be deemed relevant for inclusion in clinical implementation. The plots depicting predicted probabilities displayed a generally well-calibrated model, although with a tendency to overestimate the risk of LC. Decision-curve analyses revealed that the DES model offers optimal usability when applied to the lowest risk interval, specifically the 1/3 of patients with a risk range of 7-35\%. Patients in higher-risk intervals do not derive any additional benefit from the model and should be screened independently, regardless of the model's outcome. 

Annual low-dose CT screening is recommended in the United States for individuals aged 50 to 80 with a smoking history of 20 pack-years who are currently smoking or have quit within the past 15 years \cite{krist2021screening}. However, if we simplify these criteria to include all current and former smokers within the same age group in our population, only 54\% of our study participants would qualify for screening. Additionally, 30\% of LC patients in our study would be classified as false negatives, as they were either non-smokers or fell outside the specific age interval. This underscores the necessity for a more sophisticated model than the one currently employed in the United States. To our knowledge, only a few larger studies have attempted to predict LC based on routine blood sample analysis \cite{gould2021machine,wang2019prediction}. They used ML approaches to study extensive American cohorts. Gould et al. introduced a straightforward yet high-performing model, achieving a sensitivity of 40.1\% at a fixed specificity of 95\% and an AUC of 0.85 \cite{gould2021machine}. It outperformed the logistic regression-based PLCOm2012 model proposed by Tammemagi et al \cite{tammemaegi2014evaluation}. At a similar specificity level our presented DES model demonstrated a lower sensitivity of 24\% and an AUC of 0.77. This difference can be attributed to distinct study designs. Gould et al. conducted a case-control study with controls randomly sampled from a Cancer Registry with identical index dates. In contrast, our study is an unselected cohort of all patients examined on suspicion of LC, where cases as well as controls are expected to exhibit signs of disease. Classifying LC cases in our study is therefore more challenging but mirrors real-world conditions. A study by Wang et al. introduced a more complex model with 118 selected features, which potentially makes it challenging to implement in clinical settings \cite{wang2019prediction}. Importantly, neither of these studies included smoking status as a primary factor in LC diagnosis. In applying ICD10 codes for smoker identification, Wang et al. classified less than 2\% of the population as smokers. This proportion does not accurately represent the overall population at risk.

We investigated a population of individuals under suspicion of LC, and our dataset showed a 25\% LC incidence. It is worth noting that in the broad field of general medicine, the estimated one-year risk of LC in individuals aged over 40 is 0.30\% and 0.15\% with and without previous cancer \cite{rubin2023developing}. The significant contrast reflects a possible need for external validation prior to applying the proposed model to a lower-risk cohort in general practice.

Individuals lacking available information on smoking status were not included in the study. This absence of a clear association between smoking and the other variables made it impractical to apply any imputation techniques for this variable. Although the current model demonstrates creditable performance, its predictive capabilities might improve by having access to a more comprehensive smoking history, including information on pack-years. It would also facilitate a more direct and precise comparison with the prevailing state-of-the-art screening criteria \cite{krist2021screening,robbins2021comparative}.

We assessed the model’s performance by comparing it with the diagnoses and predictions of five pulmonologists who evaluated 200. Importantly, the comparison has certain limitations, as it cannot be directly equated with clinical practice. In clinical settings, decisions are often based on a combination of symptoms, examination findings, and medical history, including comorbidities and the progression of laboratory results.

Our proposed model can successfully predict LC using age, gender, smoking history, and a limited set of standard blood sample analyses typically conducted during the referral process. Its greatest predictive performance relates to stage I patients, who are potentially eligible for curative treatment. Although promising, these results are based on data registered at time of diagnosis. Creating a model capable of predicting LC even before the referral stage would offer significant advantages. In this study, patients were stratified according to one risk cut-off, but a two-sided cut-off could potentially have greater clinical impact by stratifying patients into e.g., low, medium and high-risk cohorts. It is also important to consider validating the model in a low-risk population or within relevant outpatient clinics. Upon comprehensive validation, this model has potential for integration in general practice in Denmark, where smoking status and laboratory analyses are routinely recorded.

\section*{Methods}
\subsection*{Study cohort}
This study retrospectively collected data from all patients in the Region of Southern Denmark referred for  examination on suspicion of LC between January 2009 and December 2018 (Fig. \ref{fig:fig1}). Initially, we considered all patients based on two classification codes AFB26 and DZ031B indicating the initiation of referral for LC diagnostics at one of the four regional LC fast-track clinics (see Supplementary Fig. S1 online). The codes were delivered from the regional data warehouse, and pertain to The Danish Medical Classification System (SKS), based on the World Health Organization’s International Classification of Diseases, currently ICD10 \cite{Sund}. To identify LC patients in the cohort, we cross-referenced it with records in the Danish Lung Cancer Registry, and incorporated 1,646 LC patients who did not follow the standard LC fast-track pathway. Due to missing information on gender and a prior history of LC, respectively, 56 and 283 patients were excluded. The final cohort included 38,944 patients, of which 11,284 were diagnosed with LC and 27,660 were not. 

\subsection*{Data collection}
Data on laboratory test results and smoking status were obtained from the regional data warehouse. The laboratory results were collected within 28 days before and 14 days after the date of the assigned SKS codes, referred to as the \textit{index date}. For patients who bypassed the LC fast-track clinic, the index date was substituted with that of the LC diagnosis registered. In case multiple index dates were available, the first index date was considered. We included the 20 most frequent blood sample analyses within the LC diagnostic clinics. Since amylase, total calcium, and INR were infrequently used by two of the clinics, their lack of the three analyses was accepted as missing and imputed. Information on smoking status was extracted from the electronic health records as free text and annotated manually by a medical doctor. Since smoking status was not directly associated with the other variables, no imputation was performed and patients without available smoking status was excluded. Of the initial pool of 38,944 patients, 9,940 had data on both a minimum of 17 laboratory results from the mentioned timespan, relevant clinics as well as smoking status. Among these individuals, 2,505 (25\%) were diagnosed with LC, and 7,435 (75\%) were found not to have LC.

\subsection*{Ethics approval}
The study was conducted in accordance with the Declaration of Helsinki (as revised in 2013) and approved by the Danish Data Protection Agency (19/30673, 06-12-2020) and the Danish Patient Safety Authority (3-3013-3132/1, 03-30-2020). Individual consent for this retrospective analysis was waived.

\subsection*{Overview of model development}
A flowchart outlining the developed ML pipeline is available in Supplementary Fig. S2. To obtain a gold standard, 2\% of the data, 200 samples, were reserved as a test set for comparison with the diagnoses made by the five pulmonologists, all experienced in evaluating patients suspected of having LC. These 200 samples were randomly selected while maintaining the overall distribution of LC and non-LC patients in the entire dataset. Given the low rate of missing values in the test set, no imputation was necessary for these 200 samples. The remaining 98\% of the data (9,740 samples) were employed for model training and validation. Hyperparameter tuning was conducted using a 2-fold cross-validation technique to identify the optimal configurations. For model training we applied a stratified 5-fold cross-validation approach to ensure that the training and validation sets maintained the same proportion of LC and non-LC cases consistent with the composition of the entire cohort. The training data underwent imputation based on median values, scaling, and class-imbalance handling via RandomUnderSampler \cite{lemaavztre2017imbalanced}. We trained four classification algorithms, specifically Logistic Regression (LR), Extreme Gradient Boosting (XGBoost), Light Gradient Boosting Machine (LGBM), and Support Vector Machine (SVM). The four models were subsequently combined into a Dynamic Ensemble Selection (DES) model, capitalizing on the strengths of each model (see Supplementary Fig. S5 and Supplementary Table S3 online). The SHAP method was applied to provide further insight into the explanations behind the predictions generated by the ML models.

\subsection*{Statistical analysis}
Summary statistics are presented as median with IQR and percentage in Table \ref{tab:tab1}. The Wilcoxon signed-rank test and the chi-squared test were used for continuous and categorical variables, respectively. The statistical significance level was adjusted by using the Bonferroni correction and set to a two-sided p-value less than 0.0002. To estimate model discrimination, we used accuracy, sensitivity, specificity, positive predictive value, and F1-score metrics reported at a default threshold of 0.5. The Receiver Operating Characteristics (ROC) curves were used to compare the Area Under the Curve (AUC) for different models, accompanied by standard deviations. Model performance was further evaluated through the Nemenyi test, with statistical significance set at a two-sided p-value less than 0.05 \cite{hollander2013nonparametric,demvsar2006statistical}. Model calibration was assessed by comparing predicted probabilities with the actual observed fraction of LC patients, and decision curve analyses were conducted to determine the clinical net benefit compared to default strategies of examining all or no patients \cite{vickers2019simple}. In subgroup analysis, we stratified by LC stage and created reduced models that included only the most important features, as determined by the SHAP analyses. All data analyses and ML model training were conducted on an in-house cloud service using Python (version 3.10).

\begin{table}[ht]
\centering
\begin{tabular}{|l|l|l|l|l|}
\hline
 & \textbf{Reference interval} & \textbf{LC (n=2,505)} & \textbf{Non-LC (n=7,435)} & \textbf{p-value} \\
\hline
\textbf{Age, years} &  & 75 (68-80) & 71 (59-79) & $<$0.0001 \\
\hline
\multicolumn{5}{|l|}{\textbf{Sex}} \\\cline{1-5}
Female & & 1,304 (52.1\%) & 3,273 (44.0\%) & \multirow{2}{*}{$<$0.0001} \\\cline{1-4}
Male & & 1,201 (47.9\%) & 4,162 (56.0\%) & \\
\hline
\multicolumn{5}{|l|}{\textbf{Smoking status}} \\\cline{1-5}
Never smoker & & 196 (7.8\%) & 2,288 (30.8\%) & \multirow{2}{*}{$<$0.0001} \\\cline{1-4}
Former/current smoker & & 2,309 (92.2\%) & 5,147 (69.2\%) & \\
\hline
\multicolumn{5}{|l|}{\textbf{Blood sample analyses}} \\\cline{1-5}
P-ALAT, U/L & Male: 10-70, Female: 10-45 & 19 (14-26) & 22 (16-31) & $<$0.0001 \\
\hline
P-Albumin, g/L & 34-45 & 42 (40-45) & 43 (41-45) &  $<$0.001\\
\hline
P-Amylase (pancreatic), U/L & 10-65 & 25 (19-34) & 25 (18-33) &  0.654 \\
\hline
P-Alkaline phosphatase, U/L & 35-105 & 81 (67-99) & 74 (62-91) &  $<$0.001\\
\hline
P-Basophils, $10^9$/L \tnote{1} & $<$0.02 & 0.05 (0.02-0.06) & 0.04 (0.02-0.06) &  $<$0.001\\
\hline
P-Bilirubin-total, $\mu$mol/L  & 5-25 & 7 (5-9) & 7 (5-10) & $<$0.001 \\
\hline
P-CRP, mg/L & $<$6 & 7.0 (2.3-22.0) & 3.4 (1.4-9.3) & $<$0.001 \\
\hline
Total calcium, mmol/L  & 2.15-2.51 & 2.38 (2.31-2.45) & 2.34 (2.28-2.41) & $<$0.001 \\
\hline
B-Eosinophils, $10^9$/L & $<$0.05 & 0.14 (0.08-0.24) & 0.17 (0.10-0.28) & $<$0.001 \\
\hline
B-Hemoglobin, mmol/L & Male: 8.3-10.5, Female: 7.3-9.5 & 8.5 (7.8-9.1) & 8.7 (8.1-9.3) & $<$0.001  \\
\hline
P-INR & $<$1.2 & 1 (0.94-1.08) & 1 (0.95-1.1) & 0.002 \\
\hline
P-Potassium, mmol/L  & 3.5-4.4 & 4.0 (3.8-4.3) & 4.0 (3.8-4.3) &  0.257\\
\hline
P-Creatinine, mmol/L & Male: 60-105, Female: 45-90 & 72 (61-87) & 76 (64-90) & $<$0.001  \\
\hline
P-LDH, U/L & 115-255 & 209 (182-246) & 192 (169-220) & $<$0.001 \\
\hline
B-Leucocytes, $10^9$/L & 3.5-8.8 & 8.80 (2.29-10.70) & 7.62 (6.20-9.38) & $<$0.001 \\
\hline
B-Lymphocytes, $10^9$/L & 1.0-4.0 & 1.79 (1.37-2.34) & 1.84 (1.4-2.37) & 0.071 \\
\hline
B-Monocytes, $10^9$/L & 0.2-0.8 & 0.73 (0.57-0.93) & 0.65 (0.51-0.83) & $<$0.001 \\
\hline
P-Sodium, mmol/L  & 137-145 & 139 (169-141) & 140 (138-142) & $<$0.001 \\
\hline
B-Neutrophils, $10^9$/L & 1.5-7.5 & 5.77 (4.52-7.42) & 4.66 (3.54-6.11) &  $<$0.001\\
\hline
B-Platelets, $10^9$/L & Male: 145-350, Female: 165-390 & 301 (243-378) & 271 (224-331) &  $<$0.001\\
\hline
\multicolumn{5}{l}{\begin{minipage}{6.7in}Data are presented in counts (\%) or medians (IQR). P-values were calculated using the Chi-squared test for categorical variables and the Wilcoxon rank-sum test for numerical variables. U/L: units pr. litre; g/L: milligrams pr. litre; $10^9$/L: count of cell type $\times$ $10^9$/L pr. litre; mmol/L: millimoles pr. litre; $\mu$mol/L: micromoles pr. litre. P-: Plasma. B-: Blood. ALAT: alanine aminotransferase; CRP: c-reactive protein; INR: international normalized ratio; LDH: lactate dehydrogenase. The number of digits reported on the blood test results reflects the number of digits provided by the laboratory.\end{minipage}}

\end{tabular}
\caption{\label{tab:tab1}Baseline characteristics of the 9,940 patients examined on suspicion of LC.}

\end{table}

\section*{Code and Data availability}
\label{sec:Dataavailability}
The dataset and code used for the analyses in this study are available to qualified researchers upon request. Please email the co-first author, Margrethe Hostgaard Bang Henriksen at \href{Margrethe.Hostgaard.Bang.Henriksen@rsyd.dk}{Margrethe.Hostgaard.Bang.Henriksen@rsyd.dk} or the corresponding author, Abdolrahman Peimankar, Ph.D., at \href{abpe@mmmi.sdu.dk}{abpe@mmmi.sdu.dk}. 


\bibliography{sample}

\section*{Acknowledgements}
This work was funded by the Region of Southern Denmark, University of Southern Denmark, the Danish Cancer Society, the Dagmar Marshall Foundation and the Beckett Foundation. The authors would like to thank Karin Larsen, Research secretary, The Department of Oncology, Lillebaelt Hospital, University Hospital of Southern Denmark, for helping in proofreading the manuscript. 

\section*{Author contributions statement}
A.P. designed the study and modelling, analyzed the results, supervised the project, and drafted the manuscript. MBH designed the study, collected the data, contributed to the results analysis from a clinical perspective and manuscript writing. RNHF and LS conducted all the analyses and contributed to the results analysis and manuscript writing. CLB, OH, LHJ, and TFH contributed to the results analysis from a clinical perspective and manuscript reviewing. UKW and AE co-supervised the project and contributed to the results analysis and manuscript reviewing. All authors had full access to all the data in the study and had the final responsibility for the decision to submit to publication. 

\section*{Competing interests}
All authors declare no competing interests.

%

\end{document}


\flushbottom
\maketitle

\thispagestyle{empty}

\section*{Supplementary Methods} \label{supp-methods}
\subsection*{Introduction to the lung cancer (LC) fast-track pathways in Denmark}
Danish medical guidelines emphasize the importance of promptly evaluating patients exhibiting respiratory symptoms persisting for more than four weeks, due to their elevated risk of developing LC \cite{TDH}. However, it's important to note that symptoms such as chronic coughing, breathlessness, and coughing up blood, while common in LC patients, can also be associated with other medical conditions. For example, among one hundred middle-aged patients presenting with such symptoms and a smoking history, only one is typically diagnosed with LC \cite{DLCG}. Furthermore, a significant portion of LC patients, approximately one-third, may not manifest any specific symptoms \cite{hamilton2009caper}. These challenges underscore the complexities of diagnosing LC in general medical practice, especially during the early stages of the disease.

In Denmark, LC patients are diagnosed through specialized LC fast-track clinics, where specific and well-defined procedures are employed, including CT scans, laboratory analyses, and bronchoscopy.1 Patients referred to these clinics receive classification codes (AFB26 and/or DZ031B) within the Danish Health Care Classification System, signifying the initiation of diagnostics or suspicion of LC \cite{Sund}. LC patients who receive a confirmed diagnosis are registered in the Danish Lung Cancer Registry with the ICD-10 code of C34, labelling bronchus and lung malignancy \cite{who}. However, some LC patients, especially those without specific symptoms or a clear suspicion of LC, may bypass the fast-track clinics and are registered in the healthcare system as LC cases without prior classification codes \cite{guldbrandt2015role}.

\subsection*{Study population and data collection}
\begin{figure}[h!]
    \centering
    \includegraphics[width=0.9\linewidth]{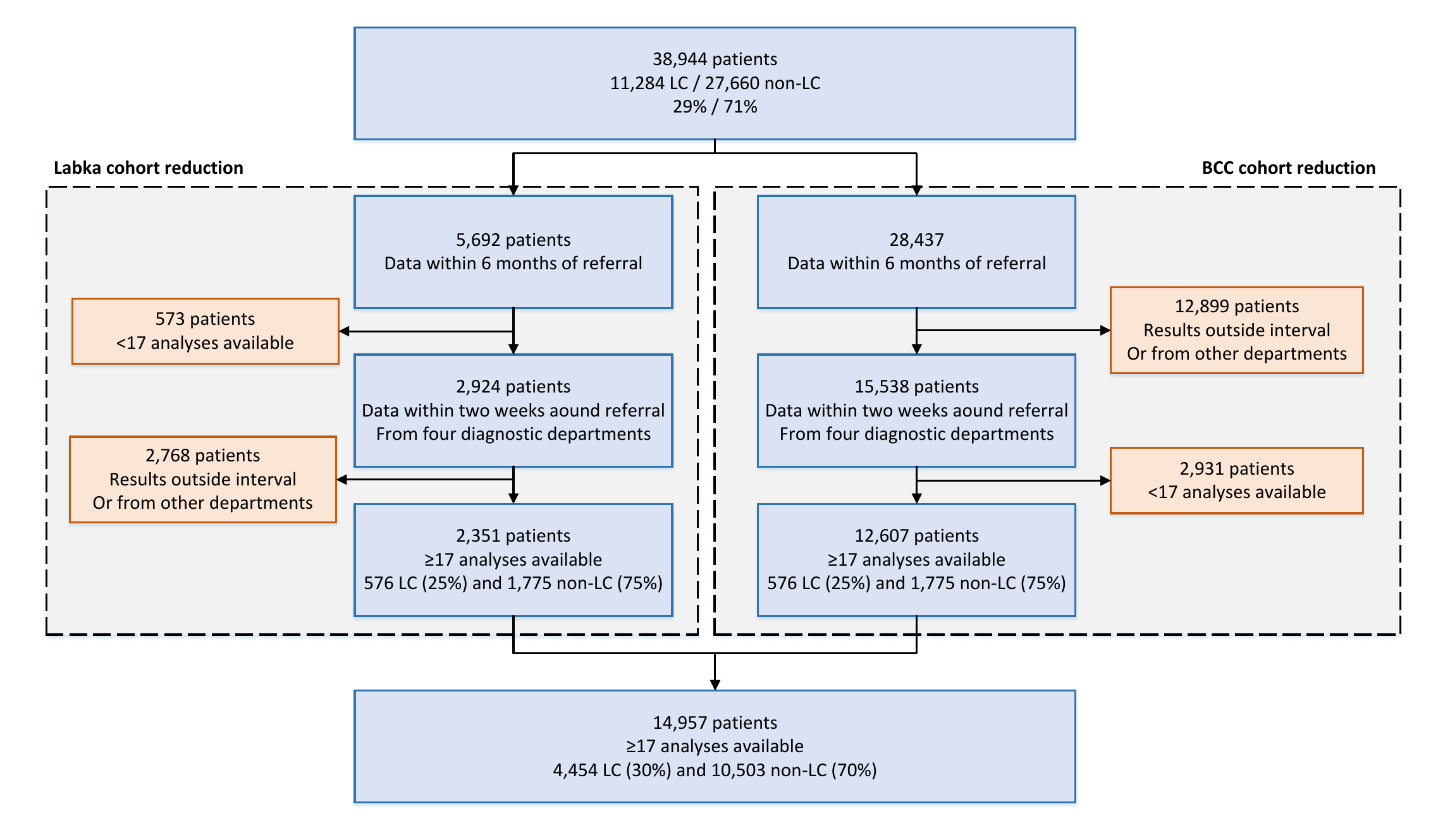}
    \caption{Cohort reduction due to relevant filtering of data. BCC: Current laboratory system utilized in the Region of Southern Denmark. Labka: The former laboratory system previously employed in specific departments within the Region of Southern Denmark.}
    \label{fig:supp-fig1}
\end{figure}

In this study, we use the date of assignment of the DZ031b or AFB26 code (referred to as the "Index Date") as the reference point for collecting blood sample analyses. Some patients in our study were referred to the LC fast-track clinic multiple times, resulting in multiple Index Dates (IDs). For consistency, we selected the first ID as the point of interest. It is worth noting that for the 1,646 patients who bypassed the LC fast-track clinics, we replaced the missing ID with the LC diagnosis date, which typically falls within the first 30 days of diagnostic initiation. We retrieved all available laboratory test results within a 180-day interval before and a 14-day interval after the ID from the regional data warehouse of southern Denmark. We further filtered the data to include results from the four departments responsible for LC diagnostics. Additionally, we refined the dataset to include the most commonly performed blood analyses in one of the hospitals in southern Denmark (Vejle University Hospital). To better align with the diagnosis process at the LC fast-track clinics, we narrowed the data to a 28-day window before and a 14-day window after the ID (Supplementary Fig. S\ref{fig:supp-fig1}).

We also investigated the frequency of missing data for each of the 20 blood sample analyses, revealing that Amylase, Calcium, and INR were infrequently tested in two of the diagnostic departments. To account for this, we allowed a maximum of three missing analyses per patient and excluded patients with a higher rate of missing data. In total, our dataset comprised 14,957 patients with data available for at least 17 blood test analyses, all conducted within four weeks of the ID and ordered by one of the diagnostic departments (Supplementary Fig. S\ref{fig:supp-fig1}).

Information regarding the smoking status of the study cohort was extracted from available electronic health records (EHR). The population was categorized into two groups: "never-smokers" and "active/former smokers." Out of the initial 14,957 patients, 5,017 lacked registered smoking status information in the EHR. Consequently, our final cohort consisted of 9,940 patients with both laboratory analyses and smoking data, comprising 2,505 LC patients compared to 7,435 non-LC patients (Supplementary Fig. S\ref{fig:supp-fig1}).

\subsection*{Flowchart of machine learning pipeline}
\begin{figure}[h!]
    \centering
    \includegraphics[width=0.9\linewidth]{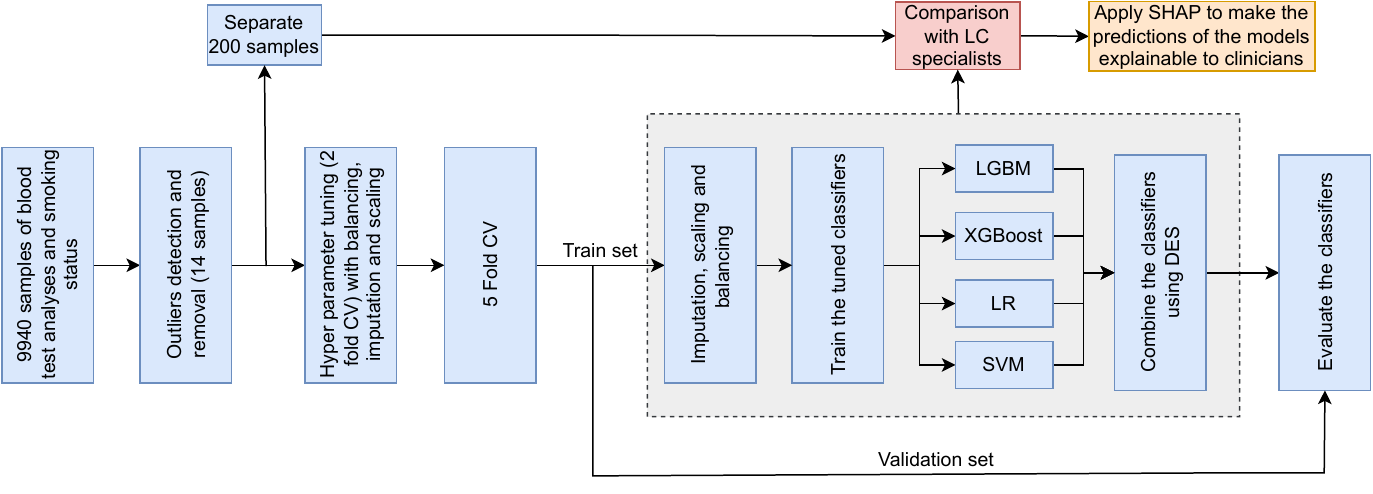}
    \caption{ML pipeline used in model preparation and training. LGBM: Light-GBM, LR: Logistic Regression, SVM: Support Vector Machine, DES: Dynamic Ensemble Selection.}
    \label{fig:supp-fig2}
\end{figure}

Supplementary Figure S\ref{fig:supp-fig2} illustrates the sequential machine learning pipeline employed in this study. In the subsequent sections, we will describe the various steps involved in this process.

\subsection*{Handling of outliers}
To ensure better generalization of the models, fourteen extreme outliers were removed from the dataset using the interquartile range technique. The range is defined as:
\[\text{Lower bound}: Q1 - 1.5\times IQR\]
\[\text{Upper bound}: Q3 - 1.5\times IQR\]

\noindent where $IQR$ represents interquartile range and $Q_1$ and $Q_3$ are the 25\textsuperscript{th}, and 75\textsuperscript{th} percentile, respectively. 
Sample falling outside the defined range were detected as outliers, which subsequently excluded from the dataset. 

\subsection*{Hyperparameter tuning}
In order to address the computational complexity and  the "curse of dimensionality" in the grid search technique, we adopted a two-stage approach to optimize the hyperparameters of the four individual machine learning models \cite{larochelle2007empirical,RICHARDBELLMAN2016}. Initially, a randomized search was used to systematically explore the parameter space, which helps determining of appropriate hyperparameter ranges \cite{bergstra2012random}. Next, a grid search technique exhaustively examined multiple combinations within the established ranges from the initial step, effectively fine-tuning all the individual models. The optimal hyperparameters for the four individual machine learning models can be found in Supplementary Table S\ref{tab:supp-tab1}.

\begin{table}[h!]
\caption{Optimal hyperparameters used for the training of the four machine learning models.}
\centering

\begin{tabular}{|l|l|}
\hline
\textbf{Model} & \textbf{Optimal hyperparameters}  \\\hline  
\textbf{LGBM} & lr=0.1; max$\_$depth=1; min$\_$data$\_$in$\_$leaf=17; min$\_$gain$\_$to$\_$split= 0; n$\_$estimators=210; num$\_$leaves=550 \\\hline
\textbf{XGBoost} & eta=0.02; max$\_$depth=3; min$\_$child$\_$weight=8; n$\_$estimators=765 \\\hline
\textbf{LR} & C=0.3; penalty=l2; solver=lbfgs \\\hline
\textbf{SVM} & C=49.1; kernel=linear \\\hline
\multicolumn{2}{l}{\begin{minipage}{6.7in}lr: learning rate; max$\_$depth: maximum depth of trees; min$\_$data$\_$in$\_$leaf: minimum number of data in a single leaf of trees; min$\_$gain$\_$to$\_$split: minimum gain to perform a split in each node of trees; n$\_$estimators: number of single trees in the LGBM; num$\_$leaves: maximum number of leaves in a single tree; eta: step size used to shrink the feature weights after each step to make the boosting process more conservative; min$\_$child$\_$weight: minimum sum of weights of the samples in a child node needed for splitting; C: inverse of regularization parameter that controls the parameters from being too large; penalty: specify the norm of the penalty term; solver: the algorithm used for the optimization; kernel: the kernel type used in the SVM algorithm.\end{minipage}}
\end{tabular} 
\label{tab:supp-tab1}
\end{table}

\subsection*{Data balancing}
Given the relatively imbalanced nature of the dataset, with 75\% non-LC patients and 25\% LC patients, we implemented data balancing techniques to prevent the model from predominantly learning the class distribution rather than the inherent characteristics of the data. We evaluated seven distinct data balancing techniques using Imbalanced-learn library \cite{JMLR:v18:16-365}. These are RandomUnderSampler, RandomOverSampler, SMOTE, BorderLineSmote, SVMSmote, KMeansSmote, and ADASYN. Our analysis of their F\textsubscript{1}-score demonstrated that most of them exhibited similar performance with overlapping standard deviations (as shown in Supplementary Fig. S\ref{fig:supp-fig3}). Ultimately, we selected RandomUnderSampler due to its downsampling approach, which efficiently reduces computational time in contrast to RandomOversampling. The RandomUnderSampler method randomly reduces the size of the majority class until a balanced distribution is achieved between the LC and non-LC classes.

\begin{figure}[h!]
    \centering
    \includegraphics[width=0.6\linewidth]{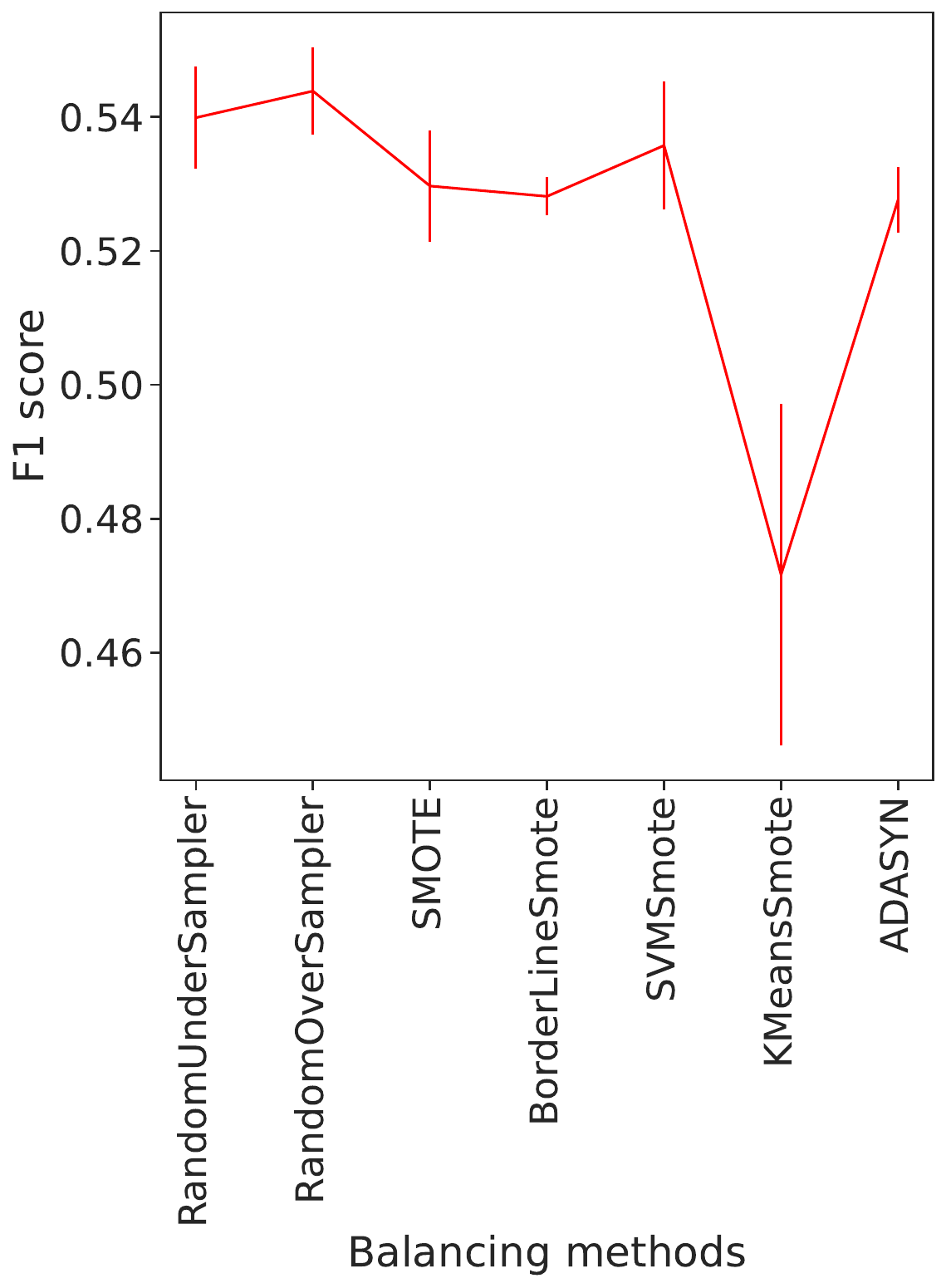}
    \caption{Comparison of different data balancing techniques applied exclusively to the LGBM model. The F\textsubscript{1}-score of different techniques along with their corresponding standard deviations are reported.}
    \label{fig:supp-fig3}
\end{figure}

\subsection*{Imputation of missing data}
We explored six different methods for handling missing data, which included mean, median, mode, k-Nearest Neighbors (kNN) imputer, iterative imputer with Bayesian ridge, and hyper-impute \cite{scikit-learn}. The mean and median imputation techniques replace missing values with the mean and median of the features, respectively. The mode imputation method fills in missing values with the most frequent value (mode) of the features. The kNN imputation method replaces missing values by first identifying the $k$ most similar samples to the sample with the missing value based on the training data set. The missing value is then imputed using the values from these k nearest neighbors. The iterative imputation method uses Bayesian ridge regression, which is a multivariate technique for imputing missing values. In this approach, a Bayesian ridge regression model is fit for each feature with missing values using other complete features as predictors. The missing values are then imputed based on the regression model's conditional predictions given the observed data. This process is repeated iteratively by re-fitting the Bayesian ridge models at each iteration until the imputed values converge \cite{jarrett2022hyperimpute}. Lastly, we experimented with HyperImpute method, which iteratively imputes missing values using an outer loop while using automatic model selection within an inner loop. The nested approach makes HyperImpute relatively computationally heavy. The features are imputed one by one, which provides the possibility to use different imputation strategies for each feature following its distribution \cite{jarrett2022hyperimpute}.

\begin{figure}[h!]
    \centering
    \includegraphics[width=0.6\linewidth]{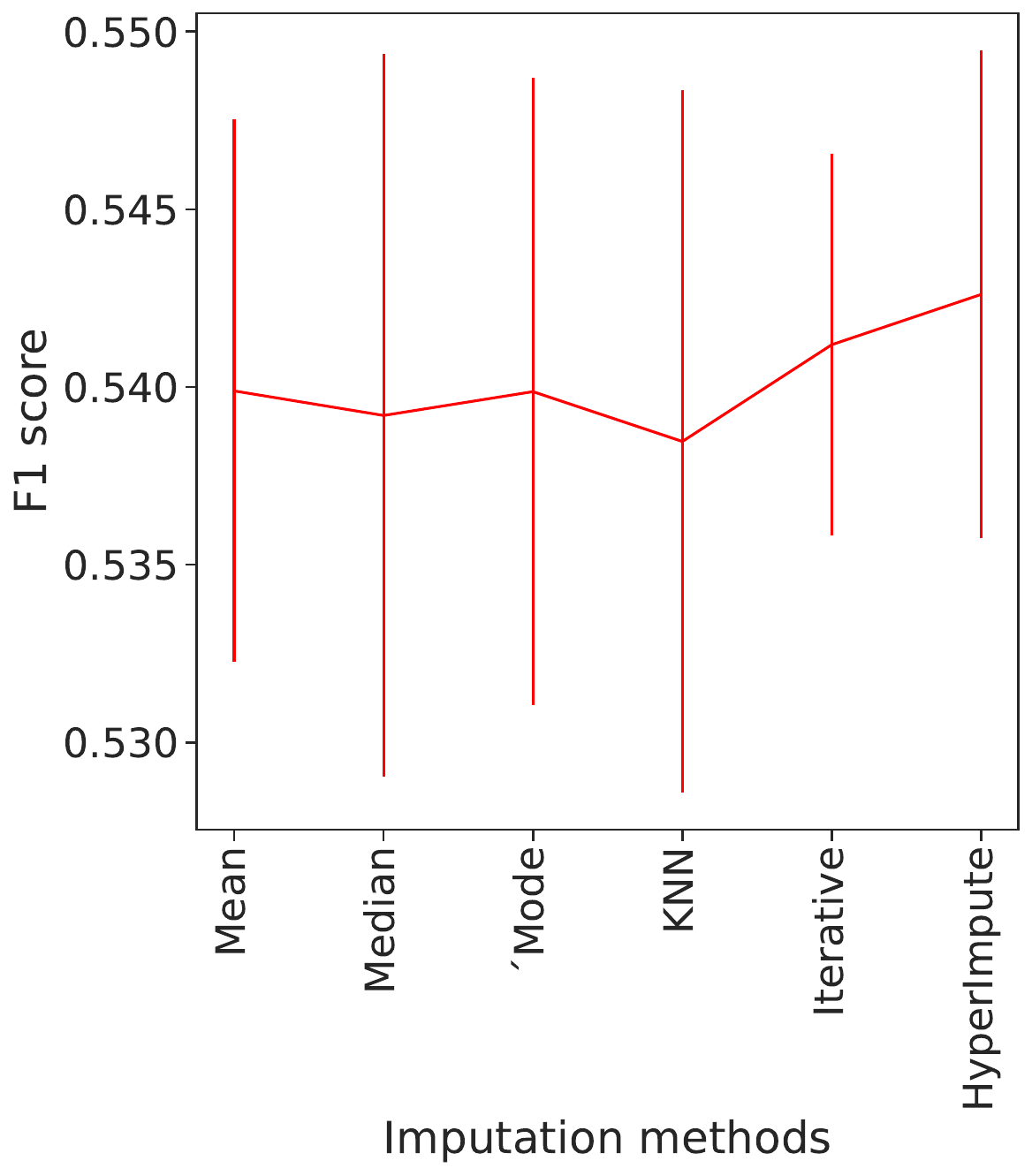}
    \caption{Comparison of different imputation techniques applied exclusively to the LGBM model. The F\textsubscript{1}-score of different techniques along with their corresponding standard deviation are reported.}
    \label{fig:supp-fig4}
\end{figure}

Supplementary Figure S\ref{fig:supp-fig4} presents the mean F\textsubscript{1}-score along with their corresponding standard deviations for these methods. Subsequent post-hoc analysis using a Friedman test revealed that none of the techniques exhibited significant differences at an alpha level of 0.05 (as detailed in Supplementary Table S\ref{tab:supp-tab2}). Given the skewed distribution observed in most laboratory variables and the simplicity of the median imputation method, we selected this approach. Additionally, imputation based on the median is a well-established strategy in the field of machine learning within health sciences. Numerous studies have demonstrated performance improvements using median imputation strategies \cite{kibria2022ensemble,berkelmans2022population,gould2021machine,rios2022handling,zhou2001multiple}. Consequently, all missing values in our dataset were replaced with the median values of the corresponding feature columns.

\begin{table}[h!]
\caption{Comparison of different imputation methods for the LGBM model using Friedman Post-hoc test. The significance level is set to 0.05.}
\centering

\begin{tabular}{|l|l|l|l|l|l|l|}
\hline
\textbf{F\textsubscript{1}-score} & \textbf{Median} & \textbf{Mean} & \textbf{Mode} & \textbf{kNN} & \textbf{Iterative} & \textbf{HyperImpute} \\\hline  
\textbf{Median} & 1.0 & 0.9 & 0.9 & 0.9 & 0.9 & 0.9 \\\hline
\textbf{Mean} & 0.9 & 1.0 & 0.9 & 0.9 & 0.9 & 0.9 \\\hline
\textbf{Mode} & 0.9 & 0.9  & 1.0 & 0.9 & 0.8 & 0.8 \\\hline
\textbf{kNN} & 0.9 & 0.9 & 0.9 & 1.0 & 0.5 & 0.5 \\\hline
\textbf{Iterative} & 0.9 & 0.9 & 0.8 & 0.5 & 1.0 & 0.9 \\\hline
\textbf{HyperImpute} & 0.9 & 0.9 & 0.8 & 0.5 & 0.9 & 1.0 \\\hline

\end{tabular} 
\label{tab:supp-tab2}
\end{table}

\subsection*{Data scaling}
As part of the preprocessing phase, we performed feature scaling to minimize the risk of overfitting arising from features with significantly larger values. Standardization was applied to ensure that features with wider distributions did not disproportionately dominate model fitting. A zero mean and unit standard deviation method was used to scale the continuous features \cite{hastie2009elements}.

\subsection*{Extreme gradient boosting }
Extreme gradient boosting (XGBoost) is an implementation of gradient boosted Decision Trees. XGBoost utilizes a gradient boosting algorithm, which is an ensemble technique \cite{chen2016xgboost}. In a boosting approach, new models/trees are built to decrease the errors made by already trained models in the classifiers/trees pool. The new models are added until there are no further improvements. It should be noted that the term gradient refers to the gradient descent algorithm used to minimize the loss once the new models are added. Then, it combines all the trained models to make the final prediction \cite{friedman2001greedy}. 

\subsection*{Light Gradient Boosting Machine}
Light Gradient Boosting Machine (LGBM) belongs to the class of boosting algorithms, which is faster and can potentially achieve higher performance compared to other boosting algorithms \cite{friedman2001greedy,schapire2003boosting}. Unlike XGBoost, which uses time-consuming presorted and histogram-based algorithms to find the optimal split of the decision stamps, LGBM uses different methods called Gradient-based One-Side Sampling and Exclusive Feature Bundling to find the optimum split value by filtering out the data instances \cite{ke2017lightgbm}.

\subsection*{Logistic Regression}
Logistic Regression (LR) is a supervised linear algorithm especially used for binary classification tasks (or multi-class classification using the one-vs-rest method). The main objective of LR is to predict the probability of an input belonging to one of the classes. During the training process, LR iteratively adjusts a fitted sigmoid shaped decision boundary, aiming to minimize the discrepancy between the predicted class probabilities and the true labels, by optimizing a loss function \cite{james2013introduction}.

\subsection*{Support Vector Machine}
Support Vector Machine (SVM) is considered as one of the most well-known statistical learning algorithms, which finds the optimum hyper-planes to classify a data set into different classes or approximate a function. Suppose that we have a data set of N inputs and targets as: $Z = \lbrace (\textbf{x}_1,t_1), (\textbf{x}_2,t_2), \ldots , (\textbf{x}_N,t_N)\rbrace$, where $\textbf{x}_n \in \rm I\!R^m$ and $t_n \in \rm I\!R$ are inputs vectors (of dimension, $m$) and targets, respectively. The SVM algorithm uses this data set to approximate the function $f(\mathbf{x})$ that maps inputs to targets, as $\sum_{n=1}^{N}(\textbf{w}^\text{T}\textbf{x}_n+b)$ , where w represents the weights vector and b is the bias term. The separating hyper-plane can be determined by both $w$ and $b$ \cite{cortes1995support}. 

\subsection*{Combine the classifiers using Dynamic Ensemble Selection}
Ensemble methods introduce several advantages over single classifiers such as improved accuracy and performance especially for complex problems. They can also reduce the risk of overfitting by balancing the trade-off between bias and variance and by using different subsets and features of the data \cite{garcia2018dynamic}. The Dynamic Ensemble Selection (DES) approach automatically selects base classifiers that achieve higher performance compared to others on $k$ nearest samples. This approach helps improving the performance of the models since different regions of the dataset might have different distributions. Therefore, the selected base classifiers perform better locally on $k$ nearest samples. 

Six DES methods were evaluated namely; 1) Overall Local Accuracy (OLA), 2) Multiple Classifier Behaviour (MCB), 3) A Priori, k-Nearest Oracle Union (KNORAU), 4) k-Nearest Oracle-Eliminate (KNORAE), and 5) Meta learning for dynamic ensemble selection (METADES) \cite{JMLRdeslib}. The ensemble method that achieved the highest performance while maintaining a low standard deviation, as shown in Supplementary Fig. S\ref{fig:supp-fig5}, was the OLA (Overall Local Accuracy) classifier. The OLA classifier analyzes the $k$ neighboring samples to evaluate the performance of each individual base classifier. Then, it selects the most accurate base classifiers that demonstrates the highest competence level for the $k$ neighboring samples \cite{woods1997combination}.

\begin{figure}[h!]
    \centering
    \includegraphics[width=0.6\linewidth]{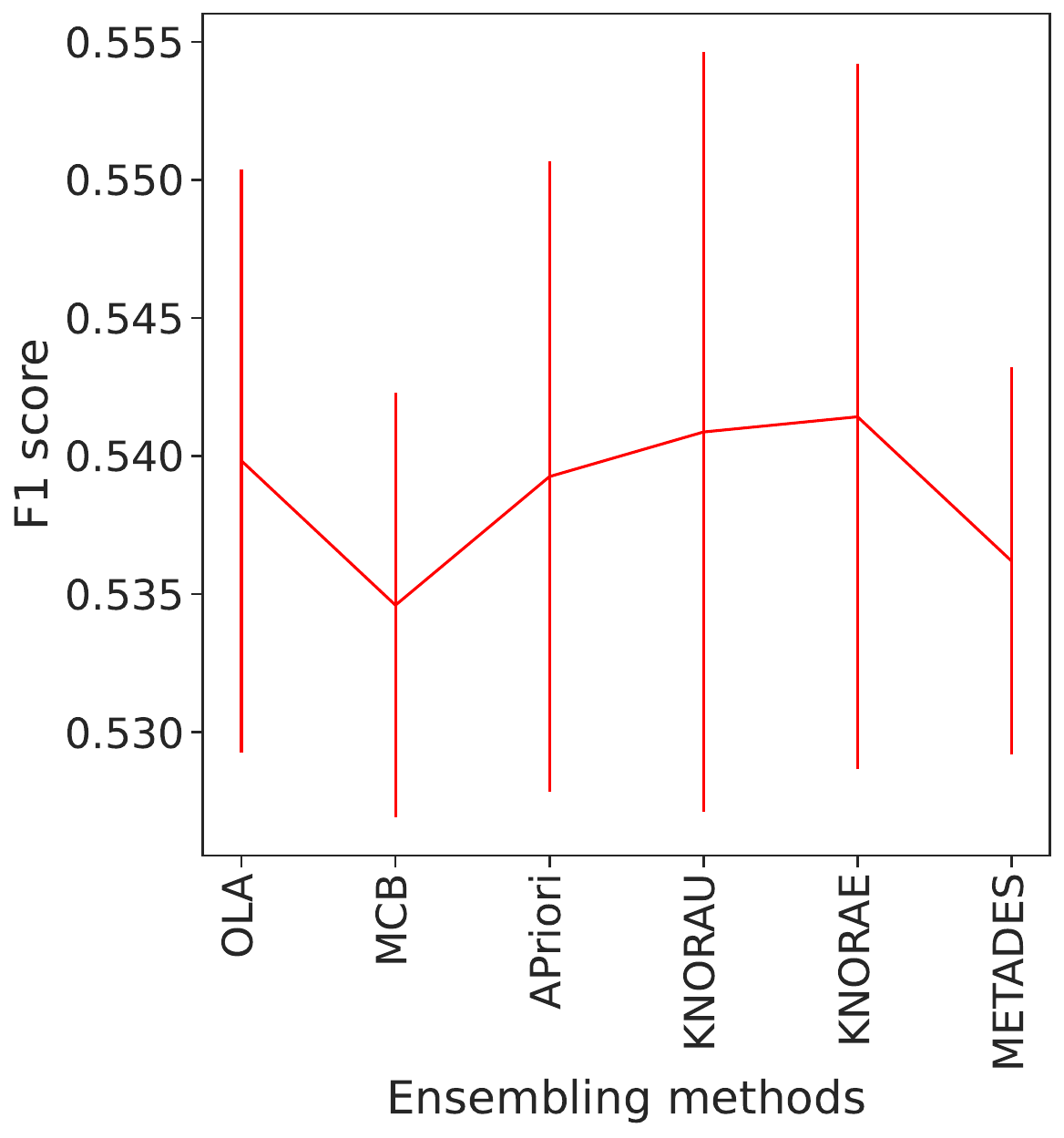}
    \caption{F\textsubscript{1}-score across the different ensemble methods along with corresponding standard deviations. OLA: Overall Local Accuracy, MCB: Multiple Classifier Behaviour, KNORAU: k-Nearest Oracle Union, KNORAE: k-Nearest Oracle-Eliminate.}
    \label{fig:supp-fig5}
\end{figure}

\section*{Supplementary Results}\label{supp-results}
\subsection*{Histograms}
\begin{figure}[h!]
    \centering
    \includegraphics[width=0.7\linewidth]{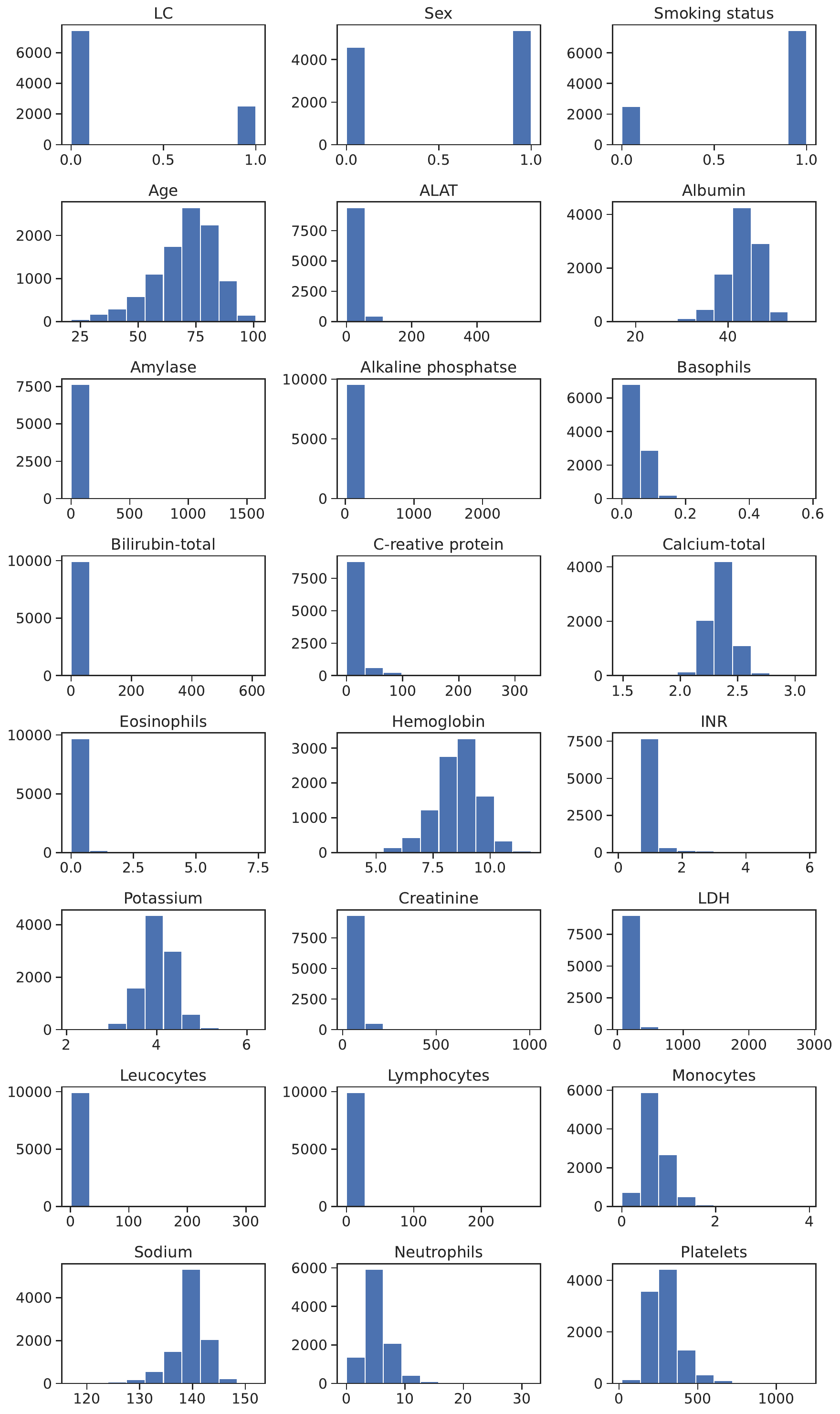}
    \caption{Histogram of available features within the dataset. The first row represents the binary label showing whether the patient has LC (lung cancer) or not, the binary features of sex, and the smoking status. The rest are the 21 continuous features including age and laboratory results.}
    \label{fig:supp-fig6}
\end{figure}

Supplementary Figure S\ref{fig:supp-fig6} shows histograms of the distributions of features used as inputs for machine learning models. It can be seen that only one of the attributes (age) has a normal distribution. Other features, such as hemoglobin and potassium have some degrees of deviations from a normal distribution. The wide and skewed histograms of some of the features indicate having outliers in the data. Furthermore, Supplementary Fig. S\ref{fig:supp-fig6} presents the class imbalance, which must be addressed before model training.

\subsection*{Boxplots}
To further examine the potential presence of outliers and to visually illustrate distinctions between the LC and non-LC groups, we also plotted boxplots in Supplementary Fig. S\ref{fig:supp-fig7}. The figure confirms that there are outliers across several features. In addition, substantial overlap can be seen between the LC and non-LC groups, particularly among the features like Albumin and Basophils. This substantial overlap shows that predicting the LC status based solely on these features can be very challenging.

\begin{figure}[h!]
    \centering
    \includegraphics[width=0.7\linewidth]{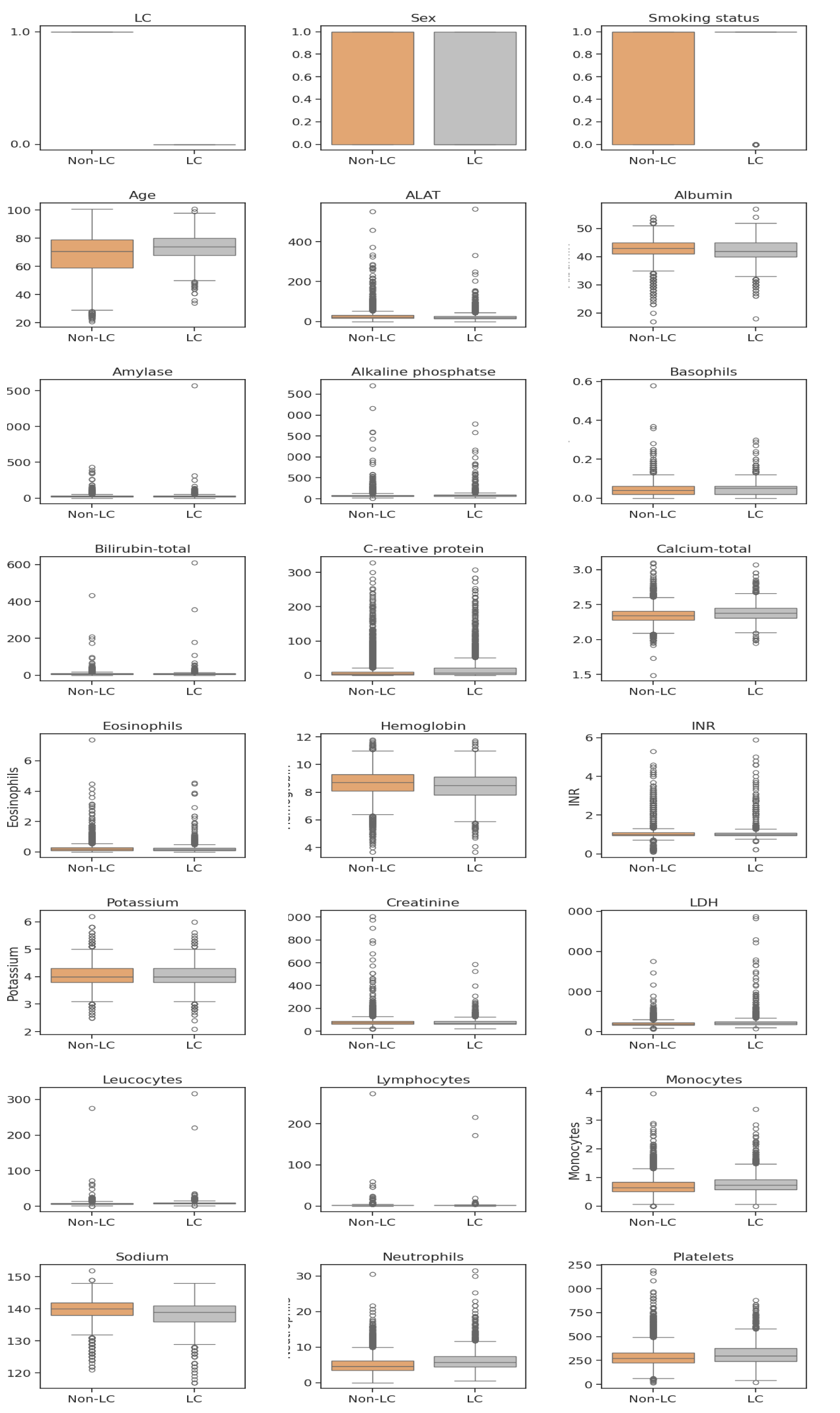}
    \caption{Boxplot of all the features in the dataset based on Lung Cancer (LC) status. While sex and smoking status are binary outcomes, the remaining variables are continuous outcomes.}
    \label{fig:supp-fig7}
\end{figure}

\subsection*{Heatmap}
\begin{figure}[h!]
    \centering
    \includegraphics[width=0.9\linewidth]{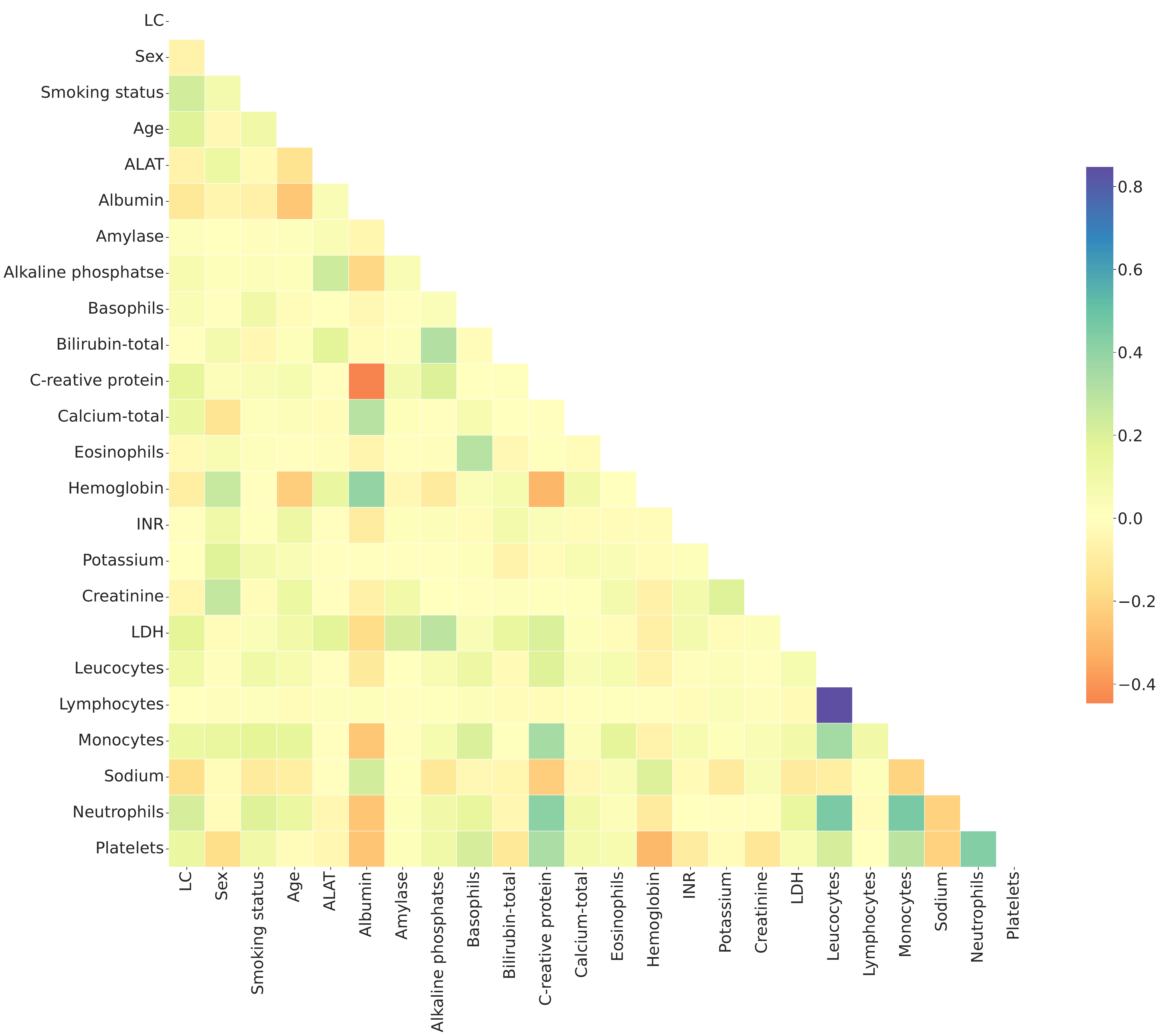}
    \caption{The heatmap illustrates the level of collinearity among the features. A high correlation value indicates a strong collinearity, while a lower index signifies minimal collinearity. As an example, Leucocytes and Lymphocytes, which represent two distinct subsets of white blood cells, show a notably high correlation.}
    \label{fig:supp-fig8}
\end{figure}

To address the issue of collinearity among features, we generated a heatmap including all the features (Supplementary Fig. S\ref{fig:supp-fig8}). The heatmap visualizes highly correlated features. For example, our analysis revealed substantial collinearity between Lymphocytes and Leucocytes, while Leucocytes and Monocytes have lesser degree of collinearity. It is important to note that some of these features naturally correlate as they measure aspects of white blood cells and their subtypes. To compare our developed model directly with the specialists, we first retained all the features in our analyses. However, we also investigated the effect of removing highly correlated features to alleviate the the risk of collinearity and overfitting.   

\subsection*{Performance evaluation}
\begin{figure}[h!]
    \centering
    \includegraphics[width=0.9\linewidth]{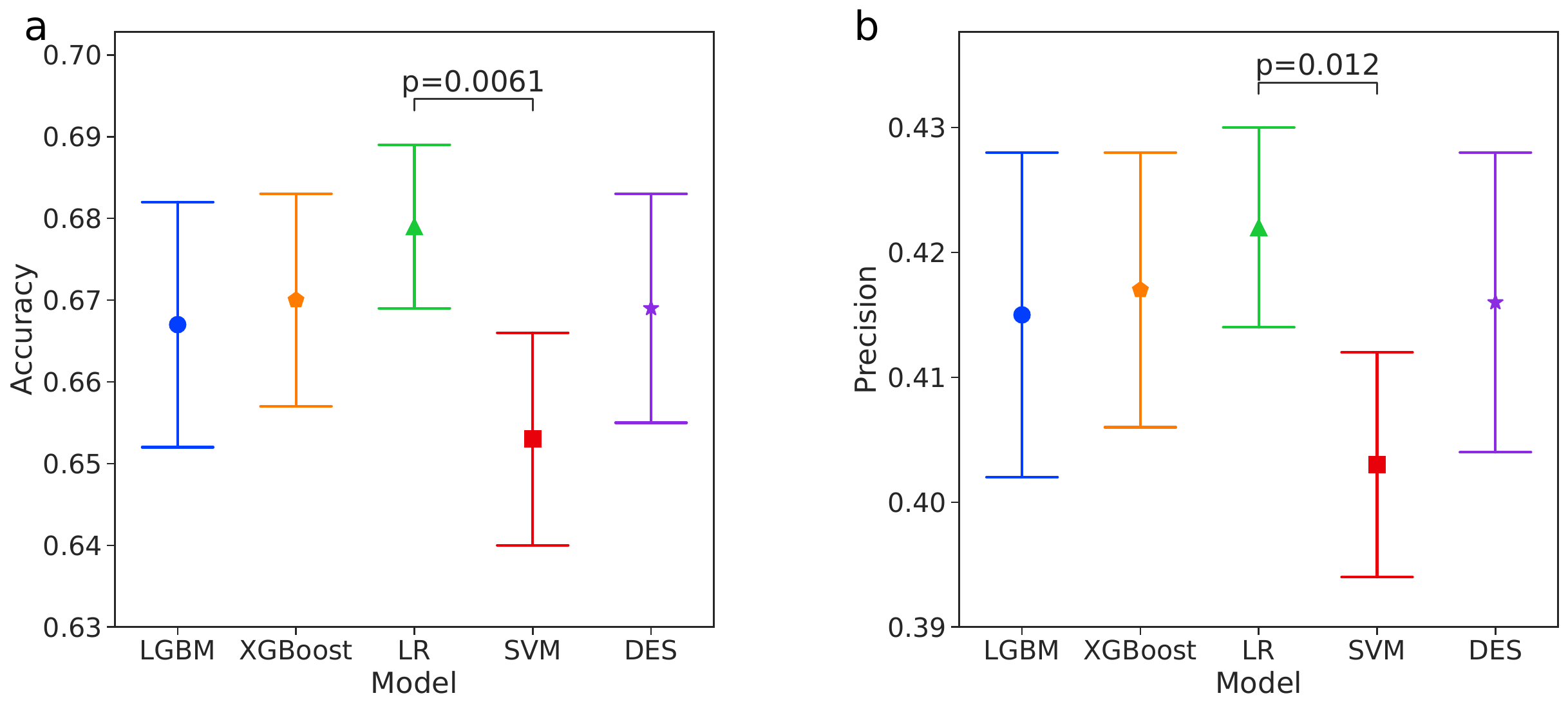}
    \caption{Comparison of different classifiers on the validation set using 5-fold cross-validation. The central marker represents mean values along with their corresponding standard deviations. The horizontal connections indicate significant differences in performance, which is determined by the Nemenyi post-hoc test with a two-sided p-value threshold of 0.05. A: Accuracy, B: Precision (Positive Predictive Value).}
    \label{fig:supp-fig9}
\end{figure}

Supplementary Table S\ref{tab:supp-tab3} presents the average and standard deviation of performance metrics across all models evaluated on the validation set using 5-fold cross-validation. The models' performances were closely aligned, and the low standard deviations suggest that the validation results are relatively consistent and reliable. Supplementary Figure S\ref{fig:supp-fig9} displays the mean and standard deviation of accuracy (A) and precision (B) for all models on the validation set using 5-fold cross-validation. On both metrics, Logistic Regression (LR) outperformed the others, although the differences with LGBM, XGBoost, and DES were not statistically significant.

\begin{table}[h!]
\caption{Comparison of classifiers’ performance on the validation set using 5-fold cross validation. Numbers represent mean values in percentages with their respective standard deviations.}
\centering

\begin{tabular}{|l|l|l|l|l|l|l|}
\hline
\textbf{Model} & \textbf{Accuracy} & \textbf{Sensitivity} & \textbf{Specificity} & \textbf{Positive Predictive Value} & \textbf{F\textsubscript{1}-score} & \textbf{ROC-AUC} \\\hline  
\textbf{LGBM} & 66.7$\pm$1.5 & 77.4$\pm$2.2 & 63.0$\pm$2.5 & 41.5$\pm$1.3 & 54.0$\pm$1.1 & 77.1$\pm$0.9 \\\hline
\textbf{XGBoost} & 66.9$\pm$1.1 & 76.4$\pm$2.2  & 63.7$\pm$2.0 & 41.6$\pm$0.9 & 53.9$\pm$0.7 & 77.0$\pm$0.9 \\\hline
\textbf{LR} & 67.9$\pm$1.0 & 73.5$\pm$1.9 & 65.9$\pm$2.0 & 42.2$\pm$0.8 & 53.6$\pm$0.2 & 75.9$\pm$0.6 \\\hline
\textbf{SVM} & 65.3$\pm$1.3 & 77.6$\pm$2.4 & 61.2$\pm$2.5 & 40.3$\pm$0.9 & 53.0$\pm$0.5 & 75.6$\pm$2.4 \\\hline
\textbf{DES} & 67.0$\pm$1.4 & 76.5$\pm$2.2 & 63.8$\pm$2.3 & 41.7$\pm$1.2 & 53.9$\pm$1.0 & 77.0$\pm$0.9 \\\hline

\end{tabular} 
\label{tab:supp-tab3}
\end{table}

\subsection*{Feature removal plot}
\begin{figure}[h!]
    \centering
    \includegraphics[width=0.6\linewidth]{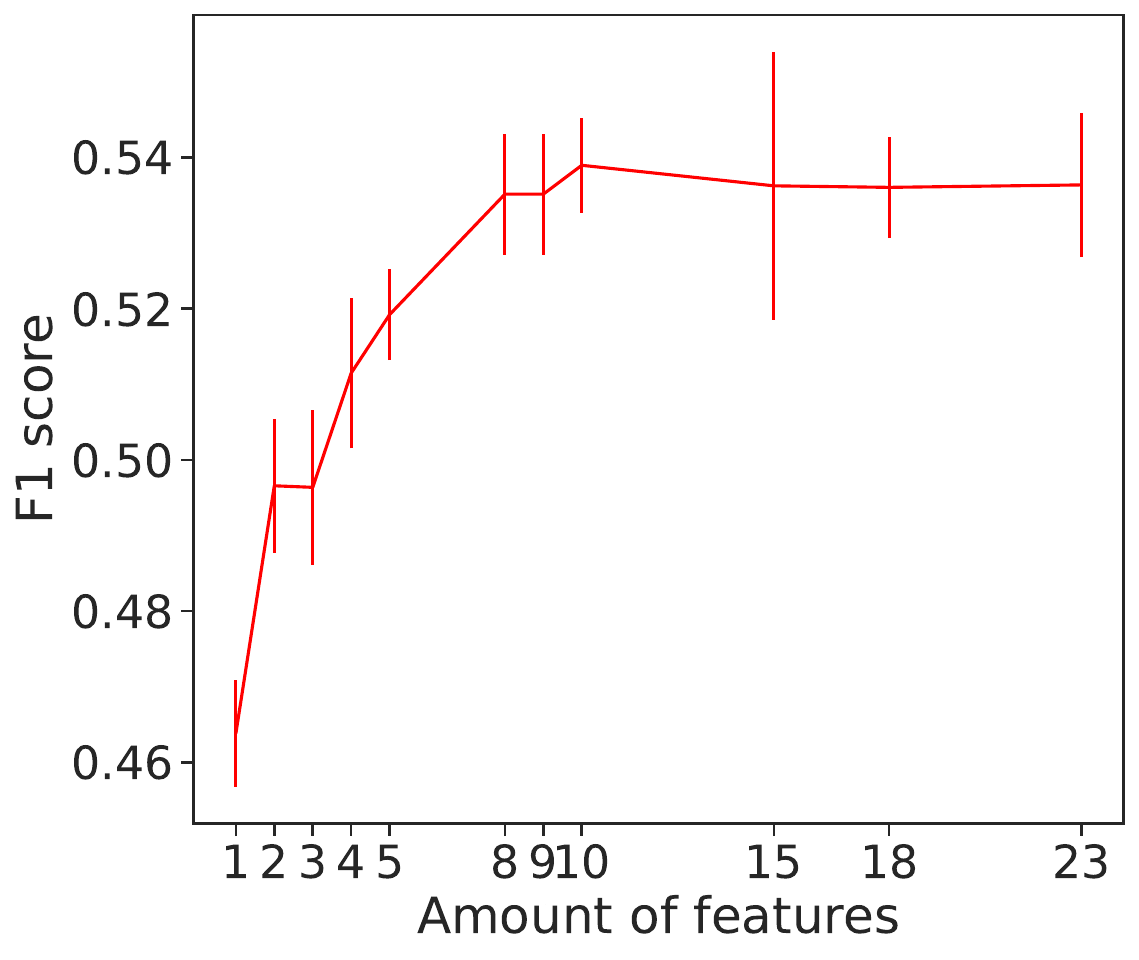}
    \caption{The effect of feature removal on the performance of the model. The horizontal lines represent the standard deviations computed from the 5-fold cross validation approach.}
    \label{fig:supp-fig10}
\end{figure}

Supplementary Figure S\ref{fig:supp-fig10} illustrates the model's performance utilizing a subset of features. In this experiment, we systematically removed features with the least impact on the final predictions as determined by their importance according to SHAP analysis. It was observed that including more than 10 features had minimal to no major effect on the model's ultimate predictions. This iterative process was conducted using a 5-fold cross-validation approach, with the standard deviations of the folds also given in the figure. Although, the DES model achieved its highest performance with 10 features, its performance using only the top five features remained comparable with only 2\% decrease in terms of F\textsubscript{1}-score.

\subsection*{SHAP individual cases plots}
\begin{figure}[h!]
    \centering
    \includegraphics[width=0.99\linewidth]{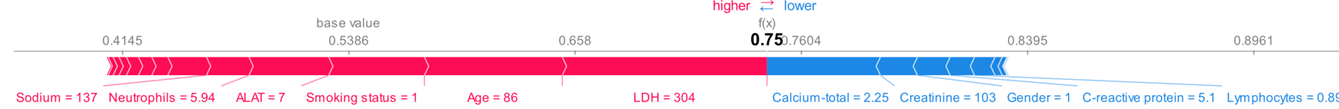}
    \caption{SHAP force plot of a true positive prediction case (i.e., an LC patient predicted as a LC patient).}
    \label{fig:supp-fig11}
\end{figure}

\begin{figure}[h!]
    \centering
    \includegraphics[width=0.99\linewidth]{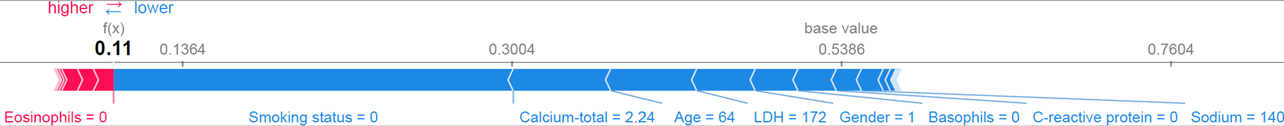}
    \caption{SHAP force plot of a true negative prediction (i.e., a non-LC patient predicted as a non-LC patient).}
    \label{fig:supp-fig12}
\end{figure}

\begin{figure}[h!]
    \centering
    \includegraphics[width=0.99\linewidth]{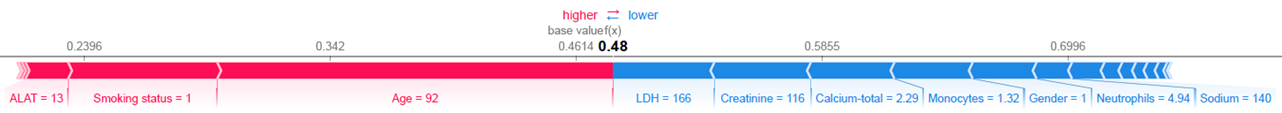}
    \caption{SHAP force plot of a false negative prediction (i.e., an LC patient predicted as a non-LC patient).}
    \label{fig:supp-fig13}
\end{figure}

\begin{figure}[h!]
    \centering
    \includegraphics[width=0.99\linewidth]{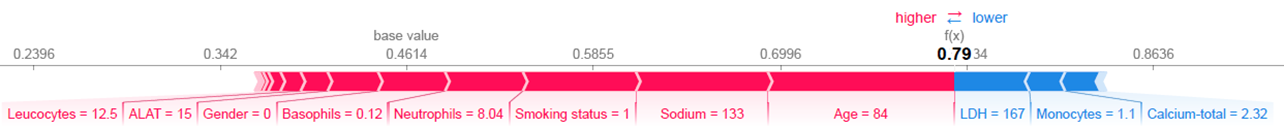}
    \caption{SHAP force plot of a false positive prediction (i.e., a non-LC patient predicted as a LC patient).}
    \label{fig:supp-fig14}
\end{figure}

To provide a more insightful understanding of the model's decision-making process for individual cases, we employed SHAP values to create Supplementary \Cref{fig:supp-fig11,fig:supp-fig12,fig:supp-fig13,fig:supp-fig14}. These are known as SHAP force plots, which help interpreting the model's predictions for each patient. Supplementary Figure S\ref{fig:supp-fig11} shows a patient with LC that has been correctly classified by the DES model with a probability of 0.75 (true positive). It is important to note that the base SHAP value (i.e., 0.5386) signifies the model's average prediction without considering any specific features. The model assigns high importance to factors like elevated LDH levels, higher age, and the patient's smoking history in its decision making process. However, the lower level of total calcium slightly pushes the plot towards a higher probability of not having LC. On the other hand, Supplementary Fig. S\ref{fig:supp-fig12} represents a correct classification of a non-LC patient (true negative), for which the DES model assigns a relatively low LC probability of 0.11. The primary contributors to such decision are being a non-smoker and lower values for total calcium, age, and LDH. Supplementary Figure S\ref{fig:supp-fig13} illustrates the outcome of an LC patient predicted incorrectly as a non-LC patient (false negative). The DES model assigns a moderately high LC probability of 0.48, which marginally misses the threshold of 0.5 for being classified as an LC case. Finally, Supplementary Fig. S\ref{fig:supp-fig14} shows the results of a non-LC patient predicted as having LC (false positive) with a relatively high LC probability of 0.79. The primary contributors to this prediction include advanced age, low sodium levels, and an active smoking status.

\subsection*{Performance on the 200 samples}
We assessed the performance of the classification algorithms by comparing their results with the diagnoses provided by five pulmonologists, utilizing a dataset comprising 200 cases. Supplementary Table S\ref{tab:supp-tab4} provides an overview of the models' performance on these 200 hold-out test cases. Notably, the LGBM model exhibited superior performance among all classifiers on these 200 cases, achieving an accuracy of 73.3\%, sensitivity of 77.2\%, positive predictive value (PPV) of 48.6\%, and an F1-score of 58.9\%. However, it should be noted that the overall performances of all models are comparable.

\begin{table}[h!]
\caption{Comparison of classification performance on the 200 test cases fixed at a specificity of 70.2\%. The numbers are in percentage.}
\centering

\begin{tabular}{|l|l|l|l|l|}
\hline
\textbf{Model} & \textbf{Accuracy} & \textbf{Sensitivity} & \textbf{Positive Predictive Value} & \textbf{F\textsubscript{1}-score} \\\hline  
\textbf{LGBM} & 73.3 & 77.2 & 48.6 & 58.9 \\\hline
\textbf{XGBoost} & 71.1 & 70.8 & 47.3 & 56.7 \\\hline
\textbf{LR} & 72.7 & 71.6 & 46.7 & 56.5 \\\hline
\textbf{SVM} & 70.0 & 70.4 & 46.7 & 56.5 \\\hline
\textbf{DES} & 73.0 & 76.0 & 47.3 & 58.3 \\\hline
\textbf{Average pulmonologist} & 69.5 & 67.3 & 42.8 & 52.3 \\\hline

\end{tabular} 
\label{tab:supp-tab4}
\end{table}

\begingroup
\renewcommand
\refname{Supplementary References}
\bibliography{sample}
\endgroup
